\documentclass[letterpaper, 10 pt, conference]{ieeeconf}  

\IEEEoverridecommandlockouts                              

\usepackage{hyperref}
\usepackage{multirow}
\usepackage{romannum}
\usepackage{graphicx}
\usepackage{amsmath}
\usepackage{amssymb}
\usepackage{breqn}
\usepackage{subcaption}
\usepackage{algorithm}
\usepackage{algorithmicx,algpseudocode}
\algrenewcommand\algorithmicindent{0.6em}

\usepackage{pgfplots}
\usepackage{booktabs, dcolumn}

\pgfplotsset{
every axis/.append style={
  axis line style={->}, 
  xlabel near ticks,
  ylabel near ticks,
  legend style={font=\scriptsize},
  label style={font=\scriptsize},
  tick label style={font=\scriptsize},
  title style={font=\scriptsize}
  }
}
\pgfplotsset{compat=newest}
\pgfplotsset{plot coordinates/math parser=false}
\newlength\figureheight
\newlength\figurewidth

\makeatletter
\let\NAT@parse\undefined
\makeatother
\usepackage[numbers,sort,compress]{natbib}
\usepackage{hyperref}
\usepackage[font=footnotesize]{caption}
\usepackage{xcolor,colortbl}

\title{\LARGE \bf
Active Perception with A Monocular Camera for Multiscopic Vision
}

\author{Weihao Yuan, Rui Fan, Michael Y. Wang, and Qifeng Chen 
\thanks{Authors are with the Hong Kong University of Science and Technology, Hong Kong SAR, China. W. Yuan (\href{mailto:weihao.yuan@connect.ust.hk}{weihao.yuan@connect.ust.hk}) and R. Fan (\href{mailto:eeruifan@ust.hk}{eeruifan@ust.hk}) are with the Department of Electronic and Computer Engineering. M. Y. Wang (\href{mailto:mywang@ust.hk}{mywang@ust.hk}) is with the Department of Mechanical and Aerospace Engineering and the Department of Electronic and Computer Engineering. Q. Chen (\href{mailto:cqf@ust.hk}{cqf@ust.hk}) is with the Department of Computer Science and Engineering and the Department of Electronic and Computer Engineering.}%
}

\begin{document}

\maketitle


\begin{abstract}


We design a multiscopic vision system that utilizes a low-cost monocular RGB camera to acquire accurate depth estimation for robotic applications. Unlike multi-view stereo with images captured at unconstrained camera poses, the proposed system actively controls a robot arm with a mounted camera to capture a sequence of images in horizontally or vertically aligned positions with the same parallax. In this system, we combine the cost volumes for stereo matching between the reference image and the surrounding images to form a fused cost volume that is robust to outliers. Experiments on the Middlebury dataset and real robot experiments show that our obtained disparity maps are more accurate than two-frame stereo matching: the average absolute error is reduced by $50.2\%$ in our experiments.

\end{abstract}


\section{INTRODUCTION}

Understanding surrounding 3-dimensional (3D) environments is an essential perception task for numerous robotic applications including manipulation, exploration, and navigation \cite{biswas2012depth, goldberg2002stereo, ye2019tightly, yuan2018reinforcement}. Robots may rely on accurate depth estimation of a scene to avoid obstacles and manipulate the objects. For depth estimation, we typically utilize depth sensors such as stereo cameras, structured-light depth sensors, and time-of-flight sensors, but depth sensors are usually expensive when compared to a single RGB camera. Researchers have been working on depth estimation with a single monocular RGB camera, but the accuracy of monocular depth is not high enough so that no popular depth sensors on the market rely on monocular depth estimation. While almost all prior works on monocular depth estimation assume passive sensing that means camera motion is uncontrollable, can we obtain accurate depth estimation with a single RGB camera combined with active sensing?

Most high-quality depth sensors are built upon the principles of stereo matching and time of flight, rather a monocular camera. Stereo cameras are equipped with two color cameras displaced horizontally so that the corresponding pixels in two cameras are on the same horizontal lines. Stereo matching estimates disparity maps that encode the differences in pixels between corresponding pixels in stereo images \cite{scharstein2002taxonomy}. Some depth sensors such as the first generation of Kinect utilize structured lights in infrared images to ease the stereo matching process but projecting infrared speckle patterns requires high power consumption. A time-of-flight sensor such as LiDAR measures the time of flight of light between the sensor and the object to further infer the depth values and also has high power consumption when emitting light. Compared to a single RGB camera, all these existing depth sensors are expensive with more cameras or projectors and have power consumption. In this paper, we show that we can obtain high-quality depth with a single camera with active sensing.


If a camera can be controlled actively (with a robotic arm), can we obtain a high-quality 3D understanding of the scene by capturing multiple images at different specified locations? With such an active sensing strategy, we have nearly perfect camera pose estimation of all the captured images and more constraints can be enforced in the reconstructed 3D model. Both the magnitude and direction of the pixels disparities can be controlled such that we can search the correspondence easily. In numerous industrial environments, a color camera is usually installed on moving agents such as autonomous ground vehicles (AGV) and robot arms that can control the camera movement. 

\begin{figure}[]
\centering
  \includegraphics[width=0.8\columnwidth, trim={0cm 0cm 0cm 0cm}, clip]{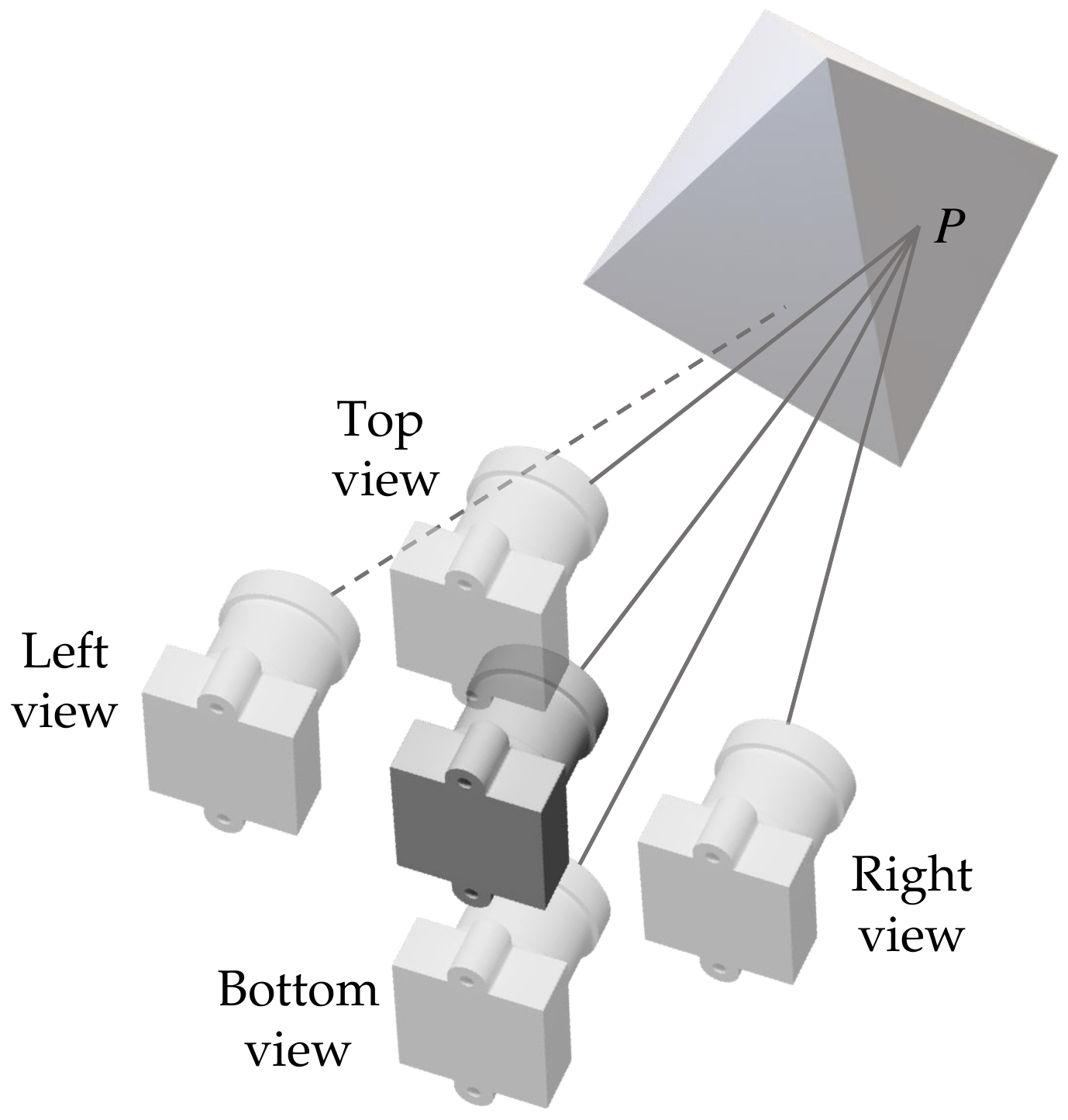}
\caption{Multiscopic vision system. The camera is moved under control in an active perception system such that the captured images are co-planar and with the same parallax. The point $P$ on a pyramid cannot be perceived from the left view but can be seen from other viewpoints.}
\label{fig:camera}
\end{figure}

In this paper, we study active perception with a single camera for depth estimation by taking multiple images at specified camera poses. We refer to the problem of depth estimation with multiple images captured at aligned camera locations as multiscopic vision, as an analog to stereo vision with two horizontally aligned images. Inspired by the principle of stereo vision that depth estimation with two perfectly aligned images is relatively easier than with two images with arbitrary unknown camera poses, we believe capturing multiple images with aligned camera locations can bring benefits in obtaining more accurate and comprehensive depth estimation.

We design an \textbf{active perception system} which uses a monocular camera mounted on the robot arm to produce a series of images surrounding a center image. These images are highly regular to form a super stereo framework, \textbf{multiscopic vision}, as is shown in Fig.~\ref{fig:camera}. We command the robot arm to move the camera along the image plane so that all images are flat co-planar. Then the search for pixel correspondence can be conducted only on a fixed line direction. If we further move the camera along the horizontal or vertical axis, the disparity will only be along the horizontal or vertical. And if the camera is moved with the same distance for every surrounding image, the disparity of each pixel relative to the center image should be the same, which is a strong regularization for computing an accurate disparity map. 



Multiscopic vision system brings clear benefits to depth estimation when compared to multiview stereo (MVS) and stereo matching. From MVS that can perform stereo matching between pairs of images \footnote{Note that simultaneous rectification of more than two images is generally impossible when their camera centers are not on one line \cite{szeliski2010computer}.}, our system can easily aggregate multiple cost volumes in our framework because all the captured images are aligned horizontally or vertically. For stereo matching methods, finding pixel correspondences is challenging because occlusion, reflection, illumination, no-texture can influence the matching easily. In a multiscopic system, depth estimation is much more robust in the presence of multiple cost volumes that can be easily combined. 

Our experiments show that multiscopic vision with multiple aligned images generates much more accurate depth estimation than stereo matching methods with only two images. Furthermore, the depth map produced by multiscopic vision contains fairly few occlusion pixels because each pixel in the central image is likely to appear in one of the surrounding images, as shown in Fig.~\ref{fig:camera}.





Our main contributions concerning monocular active perception for multiscopic vision are summarized as follows:

\begin{enumerate}
\item
We design an active perception system that captures multiscopic images with arbitrary baselines using a monocular camera mounted on an eye-in-hand robot arm.
\item
Our proposed multiscopic vision system produces more accurate depth maps by utilizing multiple images in co-planar, same-parallax structure. Also, the computed depth maps are nearly occlusion free.
\item
We evaluate and validate our multiscopic vision system on a public benchmark dataset as well as our real-world applications.
\end{enumerate}


\section{RELATED WORK}
\label{sec:related_work}


We first review prior works related to stereo vision with active perception and then camera array systems for capturing multiple images. Afterward,  we will discuss stereo vision systems with a monocular camera.

Active perception is widely employed in robotic applications such as exploration and manipulation \cite{chen2011active, bajcsy2018revisiting}. Active movement can assist in the localization of the manipulated objects under occlusions \cite{kahn2015active} or explore an unseen environment better \cite{isler2016information}.
For stereo vision, since the baseline is critical for correspondence matching, some works about actively adjusting the baseline were proposed \cite{klarquist1997adaptive, nakabo2005variable}. A linear slider was used in \cite{nakabo2005variable} to change the baseline of two stereo cameras such that the baseline could be adaptive to the distance between the camera and the environment. This enables better 3D reconstruction of different scenes.

Another approach for depth estimation is based on camera arrays in which many cameras are placed on arrays \cite{wilburn2005high, vaish2006reconstructing, maitre2008symmetric}. Thus the baseline could be changed by choosing camera pairs in different positions, and the cost volumes could be constructed with the fusion of redundant images to solve the partial invisibility problem \cite{vaish2006reconstructing, maitre2008symmetric}. The occluded surfaces for one camera could be reconstructed with the assistance of other cameras. However, building a camera array with multiple cameras is bulky and expensive. Another difficulty is the rectification of different cameras.

To take advantage of identical camera parameters, some stereo vision systems use a single camera to perform depth estimation. By analyzing the optical structure, Adelson and Wang proposed a single lens stereo system with a plenoptic camera that could produce photos from different viewpoints \cite{adelson1992single}. These captured images could be then used as stereo images for depth estimation. However, the stereo baseline is usually limited to the size of the lens aperture. Similar works using plates or mirrors to guide the light were proposed to obtain virtual stereo images. These optics design also introduces complex optical uncertainty and geometric calculation \cite{nene1998stereo, gao2004single, gluckman2002rectified, hu2017monocular}. 

None of these prior works exploit the high regularization of multiple images captured by the perception system. In contrast, we use a low-cost monocular camera to capture images in horizontally or vertically aligned camera positions. The cost volumes for stereo matching between the reference image and surrounding images can be easily combined together to form a robust cost volume for depth estimation.


\section{STEREO MATCHING BACKGROUND}
\label{sec:stereo}

We start illustrating how active perception can work for stereo matching with a monocular camera and present stereo matching algorithms related to our multiscopic vision system.

\subsection{Active Stereo Perception}

Our multiscopic vision system is capable of capturing multiple images that are combined to reconstruct the 3D scene. If we only use two of these images, the depth estimation problem would degenerate to a stereo matching problem. We begin with how a monocular camera can be applied to stereo matching problems.

\begin{figure}[]
\centering
\begin{subfigure}{0.45\columnwidth}
  \includegraphics[width=1\columnwidth, trim={2cm 1cm 4cm 1cm}, clip]{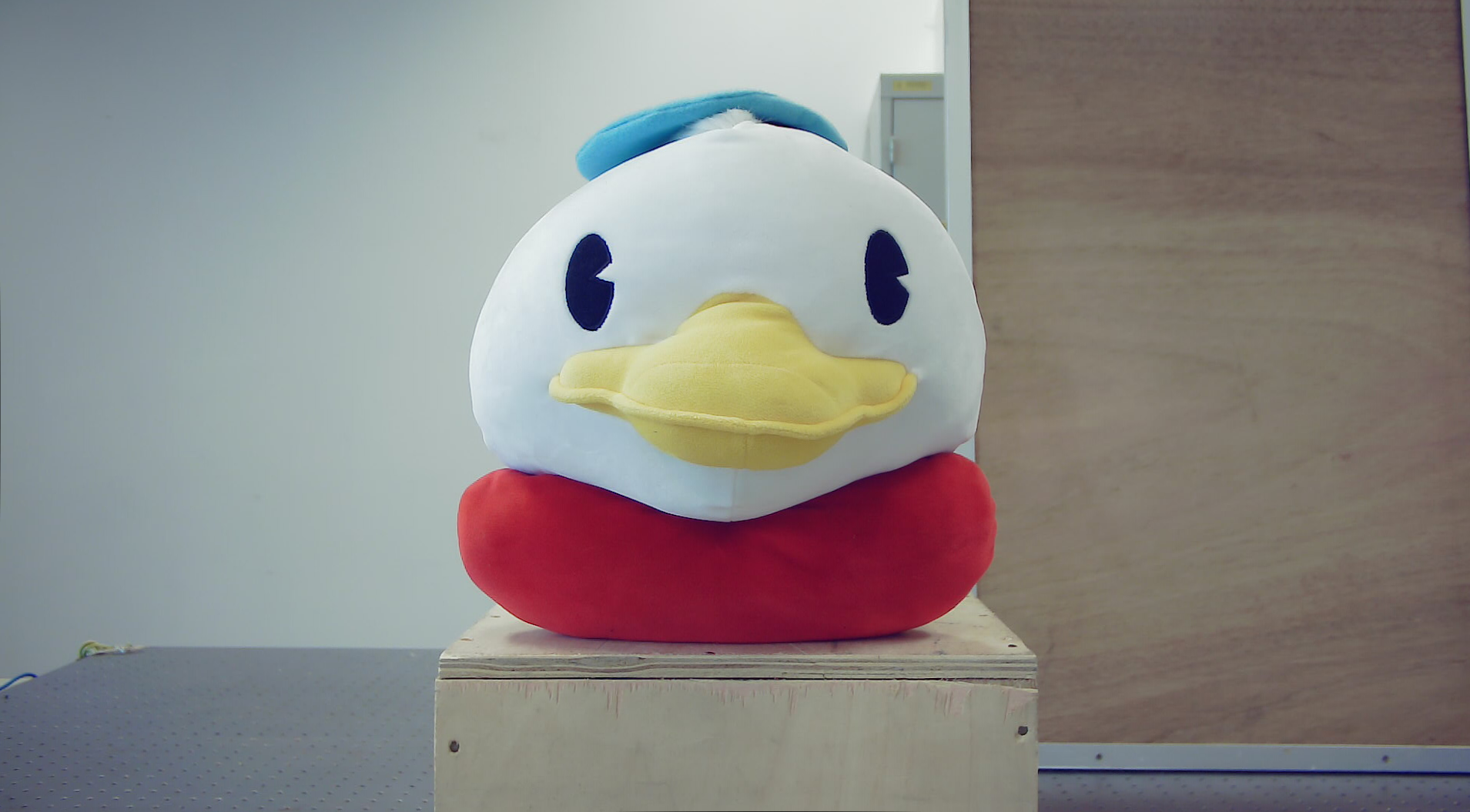}
  \caption{Left image}
  \label{fig:duck3}
\end{subfigure}
\begin{subfigure}{0.45\columnwidth}
  \includegraphics[width=1\columnwidth, trim={2cm 1cm 4cm 1cm}, clip]{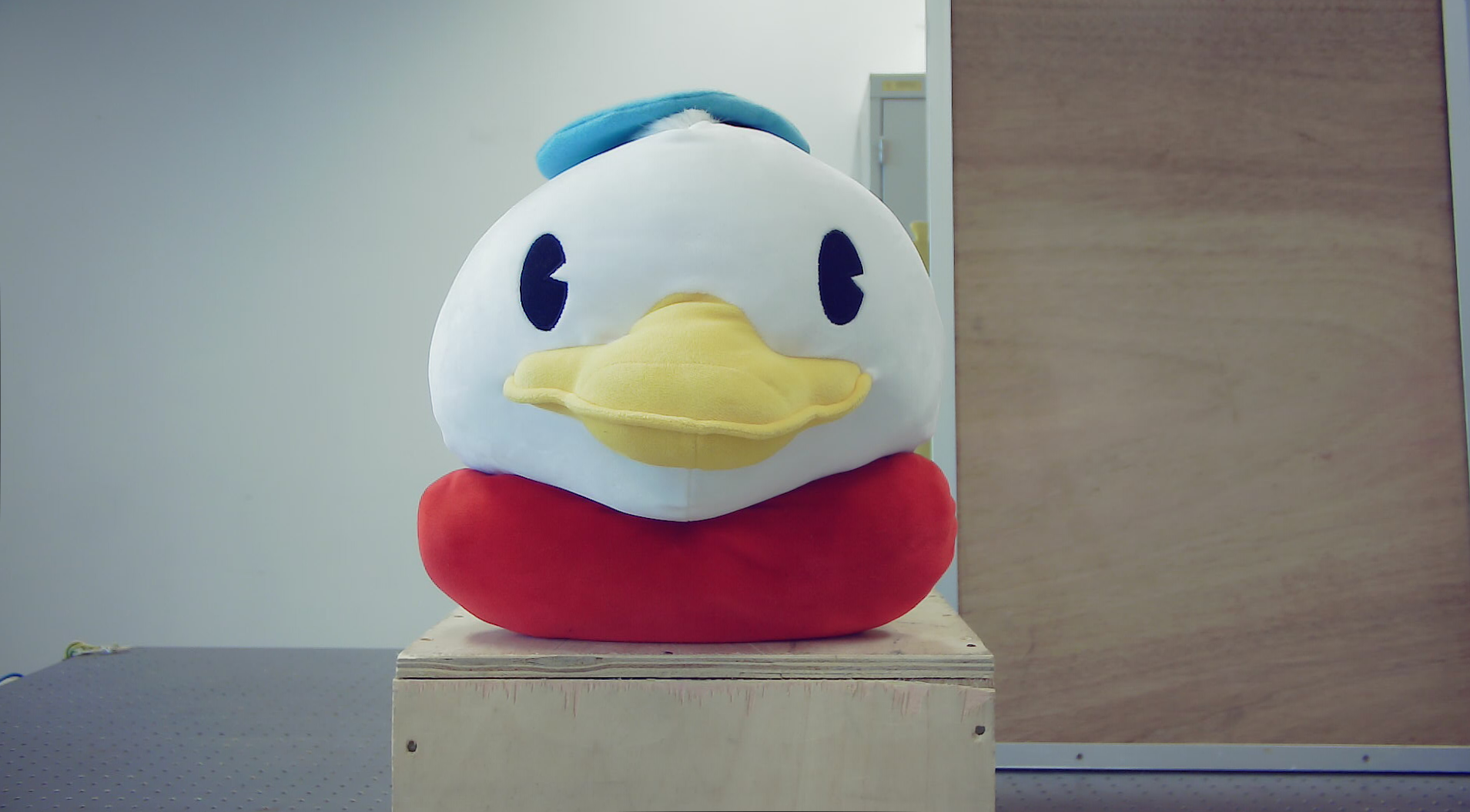}
  \caption{Right image}
  \label{fig:duck2}
\end{subfigure}
\caption{A stereo image pair captured using the proposed active perception system. The baseline between these two images is 20 mm.}
\label{fig:duck}
\vspace{-0.3cm}
\end{figure}

First, a monocular camera is installed on the end of a robot arm so the camera can be moved freely. Then we program the robot to move the camera along the horizontal axis of the image plane and two images are taken, as shown in Fig.~\ref{fig:duck}. Unlike a binocular stereo camera, the image pairs can be captured with arbitrary baselines in this active perception system. Different baselines can be applied for different purposes: accurate depth estimation of distant objects may require large baselines while stereo matching is easier with smaller baselines (less occlusion and smaller disparity).

Researchers have proposed several classic stereo matching methods such as the naive block matching, dynamic programming \cite{forstmann2004real}, semi-global matching \cite{hirschmuller2008stereo}, belief propagation \cite{sun2003stereo}, graph cuts \cite{kolmogorov2014kolmogorov} , and matching with convolutional neural networks \cite{zbontar2016stereo}. Specifically, we study the naive block matching method (BM), the graph cut method (GC), and the deep learning-based method (MC-CNN) \cite{zbontar2016stereo}.

\subsection{Block Matching}

Naive block matching is a simple and straightforward stereo matching method. This method minimizes the matching error between two blocks in the left image and the right image. To find the most similar block, we need to check all possible blocks in the same row from the minimum disparity to the maximum allowable disparity. The sum of absolute difference (SAD) is often used to measure the similarity between two blocks. For a pixel $(u, v)$ in the left image, its SAD cost with block size $2\rho+1$ and  disparity $d$ can be calculated as

\begin{equation}
  c_{\text{SAD}}(u, v, d)=\sum_{x=u-\rho}^{x=u+\rho}\sum_{y=v-\rho}^{y=v+\rho}|I_l(x,y) - I_r(x-d, y) |,
\end{equation}
where $c_{\text{SAD}}(u,v,d)$ is the cost at point $(u,v)$, $\rho$ is the radius of the block, $I_l(x,y)$ is the intensity of the pixel at $(x,y)$ in the left image and $I_r(x-d,y)$ denotes the intensity of pixel $(x-d,y)$ in the right image. The center
of the reference block is $(u, v)$ and the total number of pixels within this block is $(2\rho+1)^2$.

For the naive block matching algorithm, we simply apply the Winner-Take-All (WTA) strategy to select the correspondence with the lowest SAD cost. To improve the continuity of the results, we perform subpixel enhancement on the discrete disparity:
\begin{equation}
    d^s=d+\frac{c(u,v,d-1)-c(u,v,d+1)}
        {2c(u,v,d-1)+2c(u,v,d+1)-4c(u,v,d)},
\end{equation}
where $d$ is the integer disparity,  $d^s$ is the subpixel disparity, and $c$ is a cost volume such as $c_{\text{SAD}}$.

\begin{figure}[]
\centering
\begin{subfigure}{0.49\columnwidth}
  \includegraphics[width=1\columnwidth, trim={2cm 1cm 4cm 1cm}, clip]{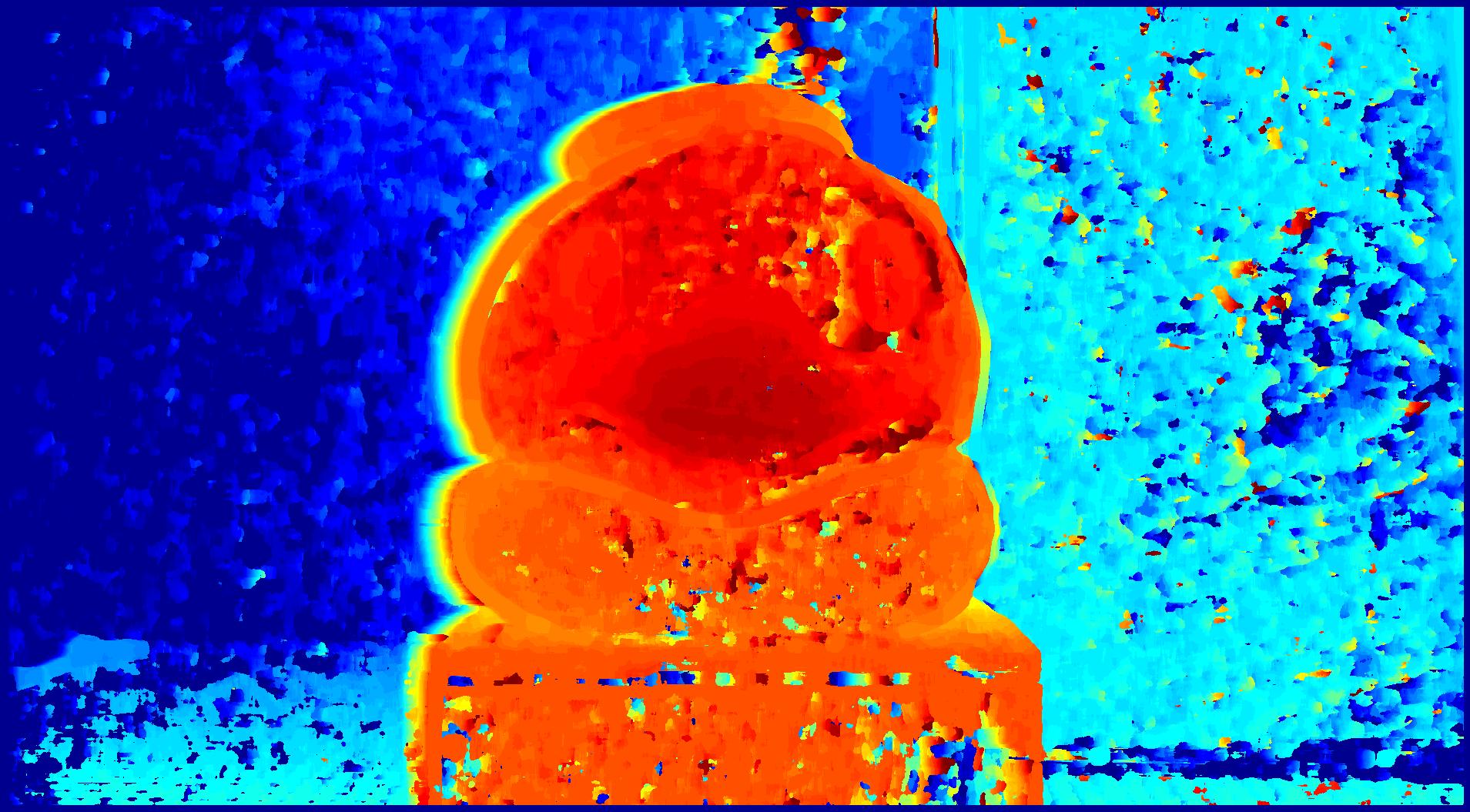}
  \caption{Block matching}
\end{subfigure}
\begin{subfigure}{0.49\columnwidth}
  \includegraphics[width=1\columnwidth, trim={2cm 1cm 4cm 1cm}, clip]{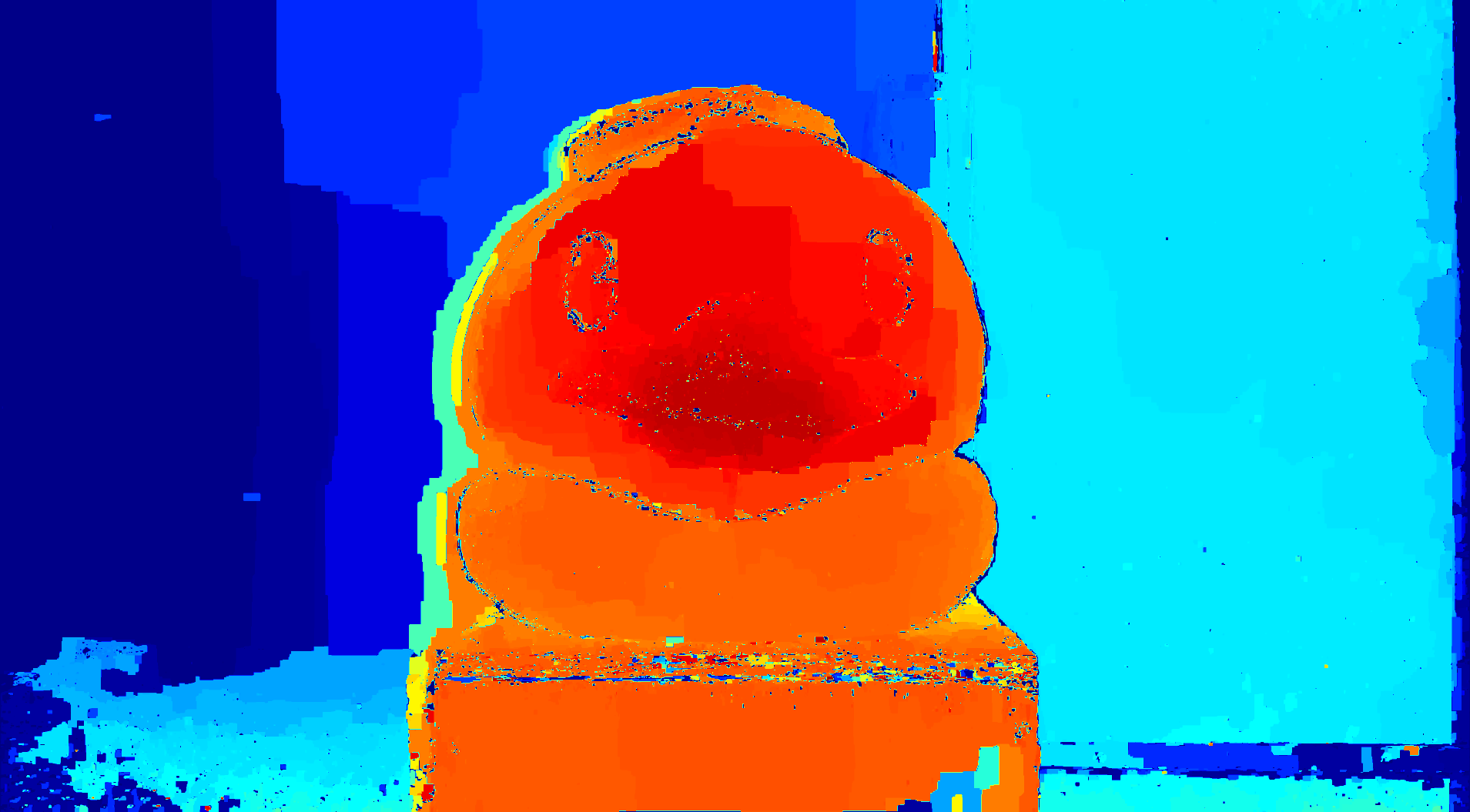}
  \caption{Graph cuts}
\end{subfigure}
\caption{The disparity maps obtained by two stereo matching algorithms, displayed in Jet colormap. (a) is using  naive stereo block matching and (b) is using stereo graph cuts. Note that the disparities on occluded regions are not estimated accurately, and the stereo matching on the metal tabletop is not accurate due to reflection.}
\label{fig:stereo}
\vspace{-0.3cm}
\end{figure}

Applying the naive block matching algorithm on image pairs captured by our active perception system, we get a noisy disparity map as displayed in Fig.~\ref{fig:stereo}(a). Also, the depths on occluded regions and reflective tabletop are not reconstructed correctly.

\subsection{Graph Cuts}

Graph cuts is one of the most popular global optimization methods for stereo matching. It is a process that assigns a label (i.e., disparity) to each pixel in the reference image such that the energy is minimized. Both our stereo graph cuts and multiscopic graph cuts are based on Graph Cuts stereo matching algorithm by Kolmogorov and Zabih \cite{kolmogorov2014kolmogorov}.

In our graph cuts optimization, the energy is composed of 4 terms defined as
\begin{equation}
  E=E_{\text{data}}+E_{\text{occlusion}}+E_{\text{smooth}}+E_{\text{unique}}.
\end{equation}

\textbf{Data term} $E_{\text{data}}$ is used to evaluate the similarity of two image patches. Note that our images may not be perfectly aligned due to the limited precision of robot arm movement, the epipolar line may deviate slightly from the horizontal or vertical direction. To compensate this, we use an improved Birchfield and Tomasi's (BT) dissimilarity for the data term \cite{birchfield1998pixel, kolmogorov2014kolmogorov}:
\begin{equation}
\begin{aligned}
  c_{\text{BT}}(u,v,d)=\max\{0, I_l(u,v)-I_r^{\min}(u-d,v), \\ I_r^{\max}(u-d,v)-I_l(u,v)\},
\end{aligned}
\end{equation}
where $I_r^{\min}$ and $I_r^{\max}$ are respectively the smallest and largest values on the subpixel neighborhood around pixel $(u-d, v)$ in the right image. For a pixel $q$ in the right image:
\begin{equation}
\begin{aligned}
  I_r^{\min}(q)=\min_\sigma\{\frac{1}{2}(I_r(q)+I_r(q+\sigma)) \},\\
  I_r^{\max}(q)=\max_\sigma\{\frac{1}{2}(I_r(q)+I_r(q+\sigma)) \},
\end{aligned}
\end{equation}
where $\sigma\in\{(0,0),(-1,0),(1,0),(0,-1),(0,1)\}$. Therefore the stereo matching for correspondence is actually performed between the half higher row and the half lower row.

\textbf{Occlusion term} $E_{\text{occlusion}}$ is used to maximize the number of matches. To encourage the disparity assignment in graph cuts optimization, any inactive pixel without assignment is penalized by energy $K$.

\textbf{Smoothness term} $E_{\text{smooth}}$ encourages to assign same disparity to adjacent pixels, especially for those with similar color. Thus if two adjacent pixels $p_1, p_2$ in left image have different disparity assignments corresponding to pixels $q_1, q_2$ in right image, a $L_1$ penalty would be added as:
\begin{equation}
\begin{split}
  V= &
  \begin{cases}
    \lambda_1 \Delta d \quad \max(|I_l(p_1)-I_l(p_2)|,\\
    \qquad \qquad \qquad
    |I_r(q_1)-I_r(q_2)|)<\theta \\
    \lambda_2 \Delta d \quad \text{otherwise}
  \end{cases}\\
  \Delta d &=\min\{|d_1-d_2|,\ d_{\text{CUTOFF}}\},
\end{split}
\end{equation}
where $\theta$ is a threshold to evaluate the color similarity, $\lambda_1, \lambda_2$ are penalty constants for similar and various pixels, $\Delta d$ is the disparity difference truncated at a threshold $d_{\text{CUTOFF}}$.

\textbf{Uniqueness term} $E_{\text{unique}}$ enforces the uniqueness of pixel correspondences. In other words, for a pixel in the left image, we do not allow two pixels in the right image match it simultaneously. This will be punished by an infinity energy $\infty$.

We use the graph cuts optimization to minimize the energy $E$. To suppress the discontinuous disparity artifacts, input images are enlarged 4 times before the graph cuts optimization. As shown in Fig.~\ref{fig:stereo}(b), the resulted disparity map with graph cuts contains less noise, but artifacts on occlusion regions and the reflective tabletop still persist.


\begin{figure}[]
\centering
\includegraphics[width=1.0\columnwidth, trim={0cm 0cm 0cm 0cm}, clip]{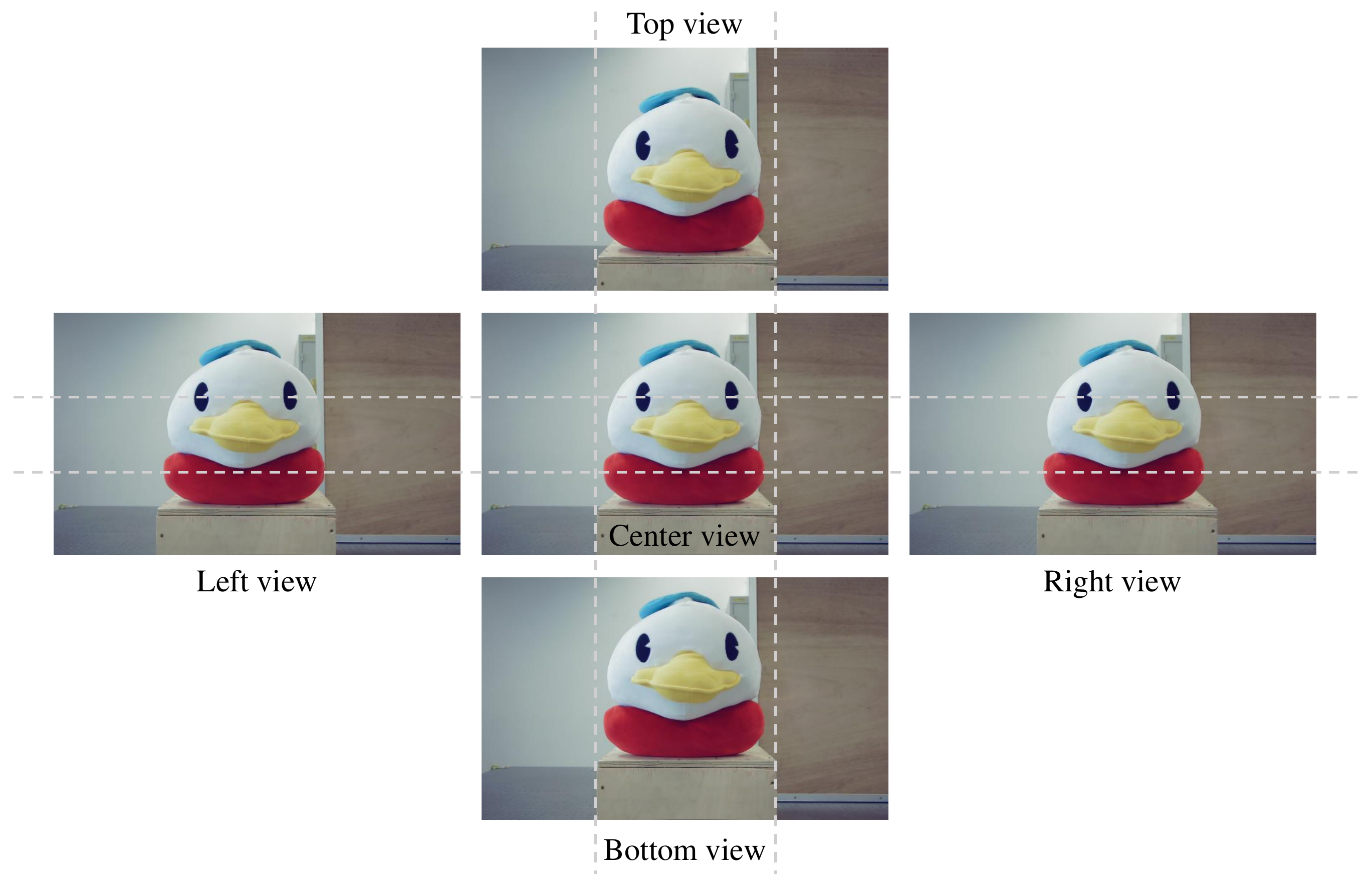}
\caption{Five images captured using our active perception system from different viewpoints. The parallaxes between adjacent views are same.}
\label{fig:5frame}
\end{figure}

\section{MULTISCOPIC VISION}
\label{sec:multiscopic}

We first introduce our multiscopic vision system with active perception to capture axis-aligned images and then propose multiscopic matching algorithms for robust depth estimation.

\subsection{Multiscopic Active Perception}


In our multiscopic vision system presented in Fig.~\ref{fig:camera}, we can move a monocular camera to the left and to the right along the horizontal axis, and move the camera up and down along the vertical axis.  We capture one center image and four axis-aligned images with the same baseline in the left, right, bottom, and top views, as displayed in Fig.~\ref{fig:5frame}. The baseline between the center image and one neighboring image is 20 millimeters. With the center image as the reference, the other four images can jointly contribute to the disparity estimation. Besides, for each point seen in the center image, it is very likely that one of the other four images would contain the point. For example, the point $P$ in Fig.~\ref{fig:camera} cannot be observed from the left view but can be perceived completely from other views.

\subsection{Multiscopic Block Matching}

\begin{figure}[]
\centering
  \includegraphics[width=0.7\columnwidth, trim={0cm 0cm 0cm 0cm}, clip]{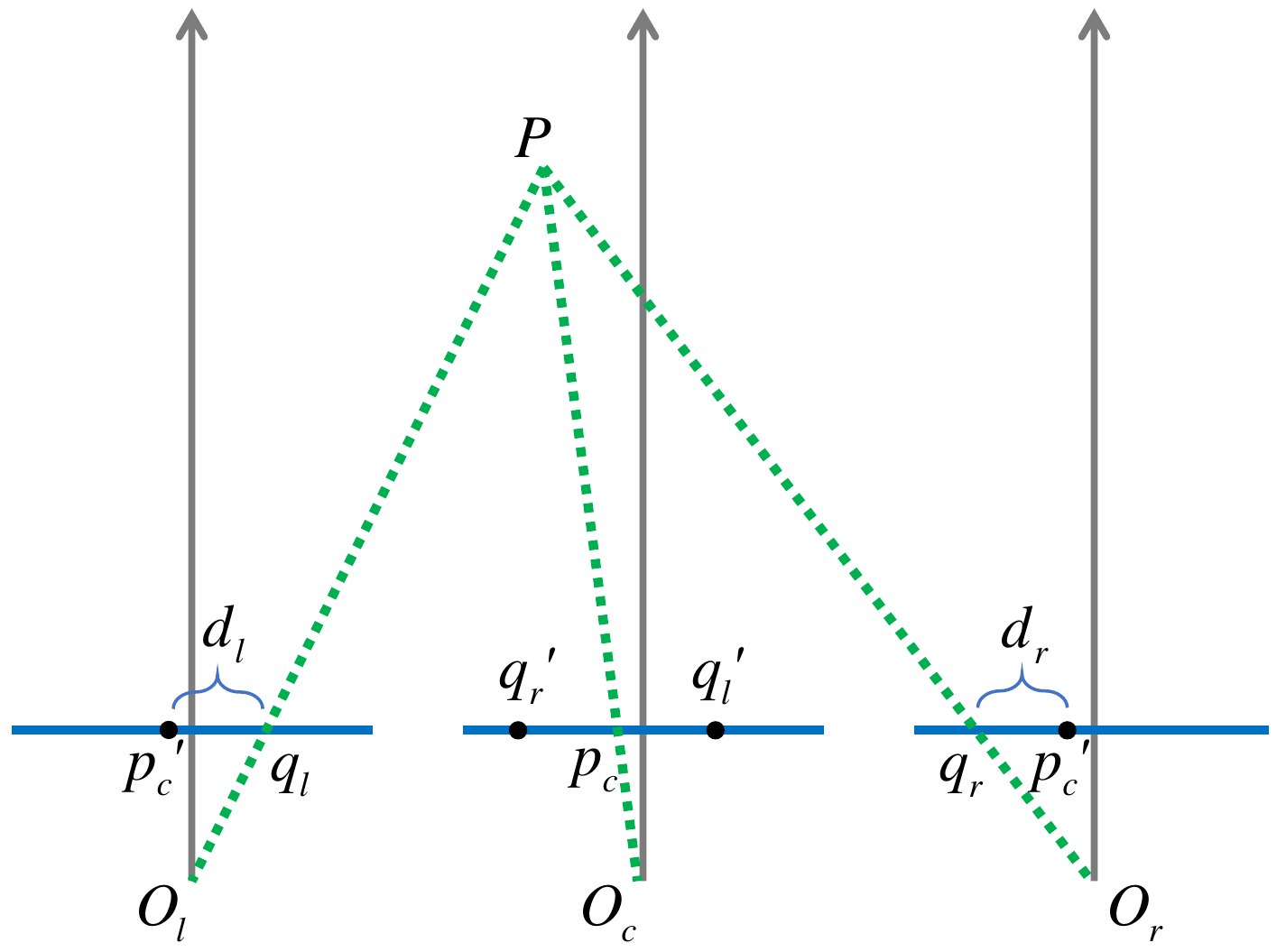}
\caption{A multiscopic structure with three images is formed by moving a camera horizontally along the image plane with the same distance. Thus there are three images captured from the left view, the center view, and the right view. The gray optical axes are perpendicular to the image planes in blue. The points $O$ are optical centers. A point $P$ in 3D space is projected onto the image plane at different time corresponding to three pixels $q_l,p_c,q_r$ in 2D images.}
\label{fig:optical}
\end{figure}

The images in multiscopic vision are taken with parallel optical axes and co-planar image planes, as illustrated in Fig.~\ref{fig:optical}. Since the baselines for four surrounding images are the same, the disparity of a pixel between the center image and any surrounding image should be the same. Note that the correspondences between the center image and another image are on the same row or column due to the horizontal or vertical movement of the camera. This is demonstrated in Fig.~\ref{fig:optical}. Considering a multiscopic system with three images as an example, for a point $P$ in 3D space, it is projected onto the camera image planes as three image pixels $q_l,\ p_c,\ q_r$. The disparity $d_l$ between $p_c$ and $q_l$ and the disparity $d_r$ between $p_c$ and $q_r$ are the same.

In the real-world application, our multiscopic vision system takes five images, as shown in Fig.~\ref{fig:5frame}. The data term is composed of four parts, each for one surrounding image: 
\begin{equation}
  \begin{aligned}
  c_{\text{SAD1}}(u,v,d)=\sum_{x=u-\rho}^{x=u+\rho}\sum_{y=v-\rho}^{y=v+\rho}|I_r(x-d, y)-I_c(x,y) |, \\
  c_{\text{SAD2}}(u,v,d)=\sum_{x=u-\rho}^{x=u+\rho}\sum_{y=v-\rho}^{y=v+\rho}|I_l(x+d, y)-I_c(x,y) |, \\
  c_{\text{SAD3}}(u,v,d)=\sum_{x=u-\rho}^{x=u+\rho}\sum_{y=v-\rho}^{y=v+\rho}|I_t(x, y+d)-I_c(x,y) |, \\
  c_{\text{SAD4}}(u,v,d)=\sum_{x=u-\rho}^{x=u+\rho}\sum_{y=v-\rho}^{y=v+\rho}|I_b(x, y-d)-I_c(x,y) |,
  \end{aligned}
\end{equation}
where $I_r,I_l,I_t,I_b$ denote the images taken from the right, left, top, and bottom views respectively.

Then fusing these four parts to form the final data term is crucial. One naive idea is to take the average,
\begin{equation}
  c_{\text{ave}}(u,v,d)=\frac{1}{4}(c_{\text{SAD1}}+c_{\text{SAD2}}+c_{\text{SAD3}}+c_{\text{SAD4}}).
\end{equation}

\begin{figure}[]
\centering
\begin{subfigure}{0.49\columnwidth}
  \centering
  \includegraphics[width=1\columnwidth, trim={2cm 1cm 4cm 1cm}, clip]{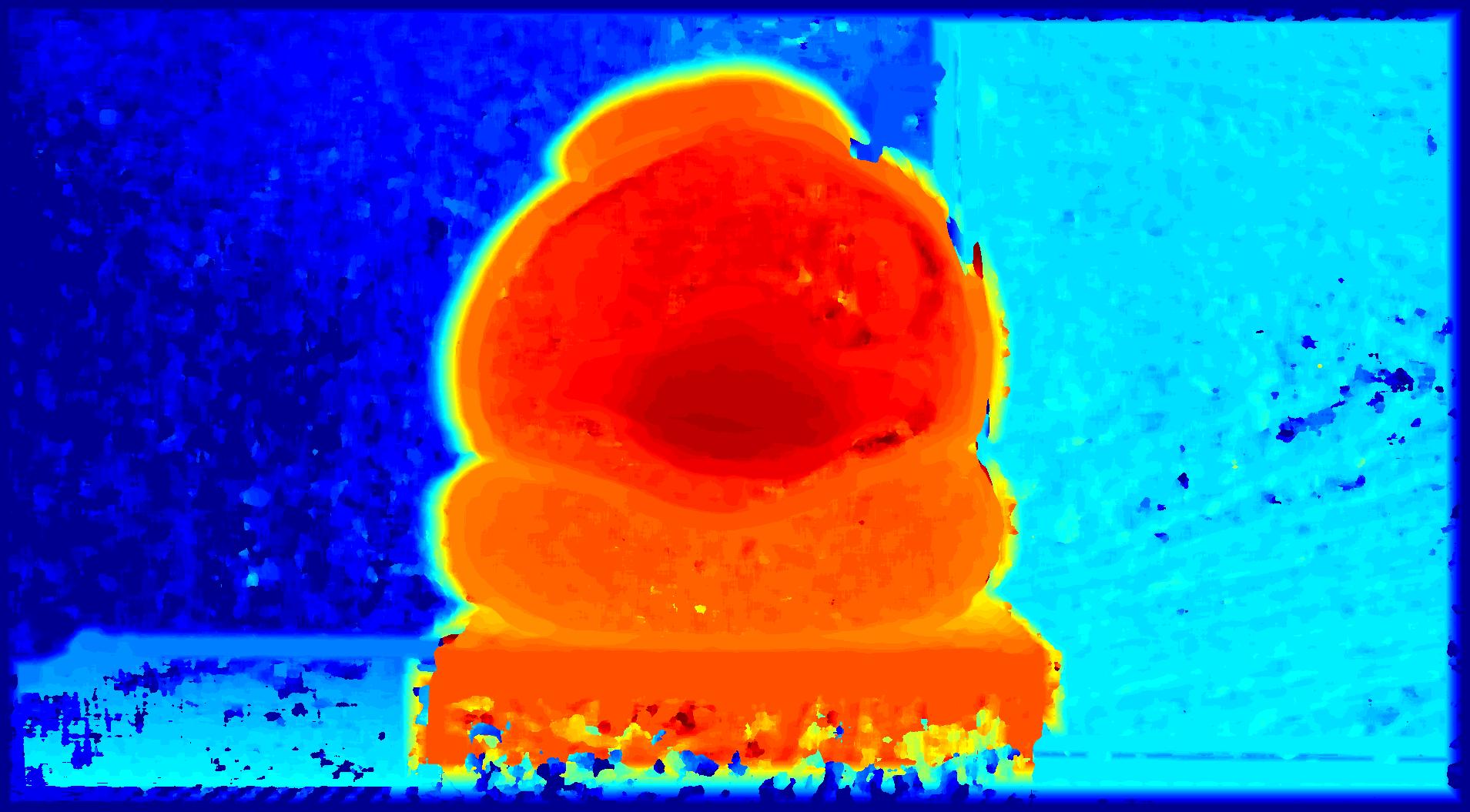}
  \caption{Mean fusion}
\end{subfigure}
\begin{subfigure}{0.49\columnwidth}
  \centering
  \includegraphics[width=1\columnwidth, trim={2cm 1cm 4Cm 1cm}, clip]{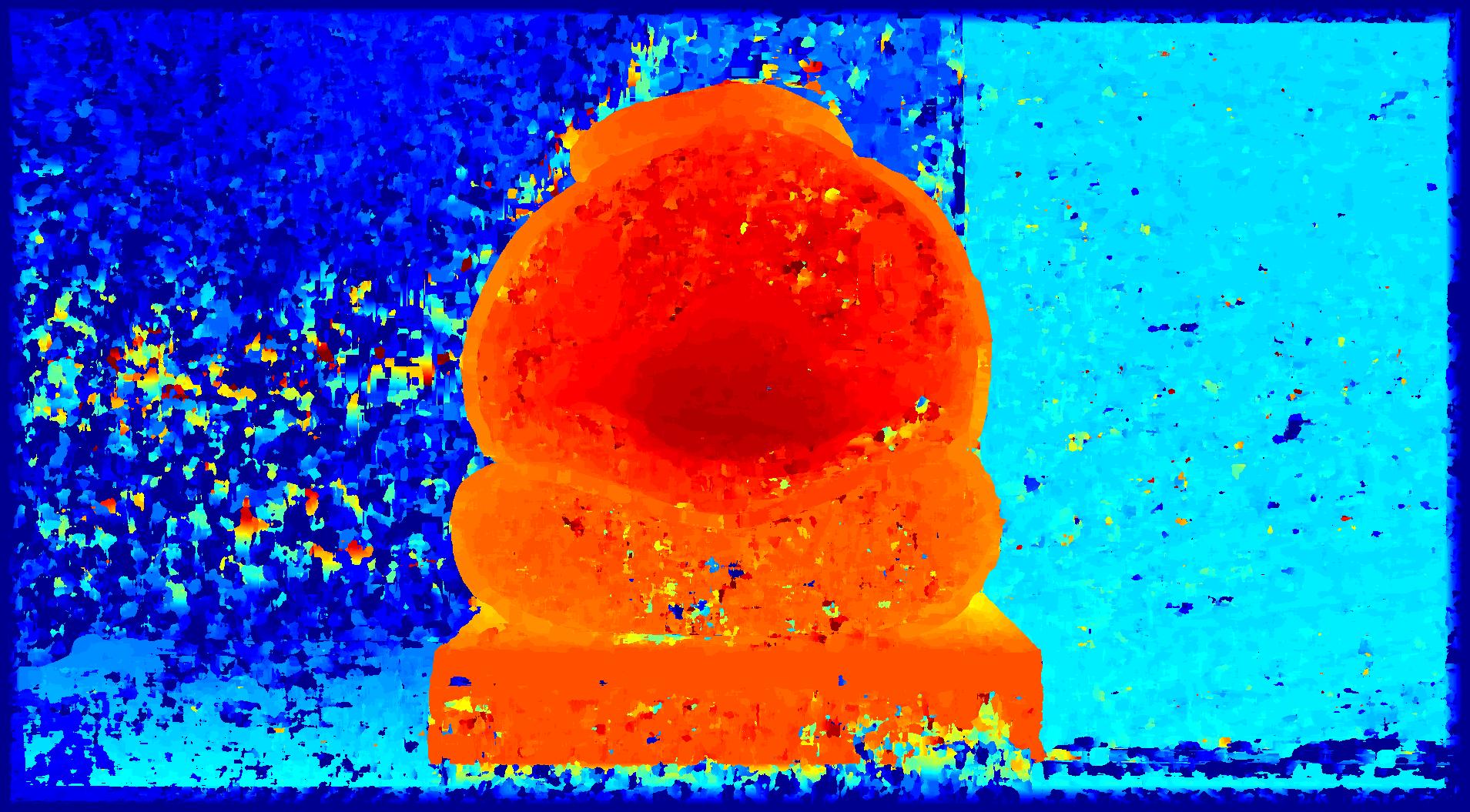}
  \caption{Minimum fusion}
\end{subfigure}
\begin{subfigure}{0.49\columnwidth}
  \centering
  \includegraphics[width=1\columnwidth, trim={2cm 1cm 4cm 1cm}, clip]{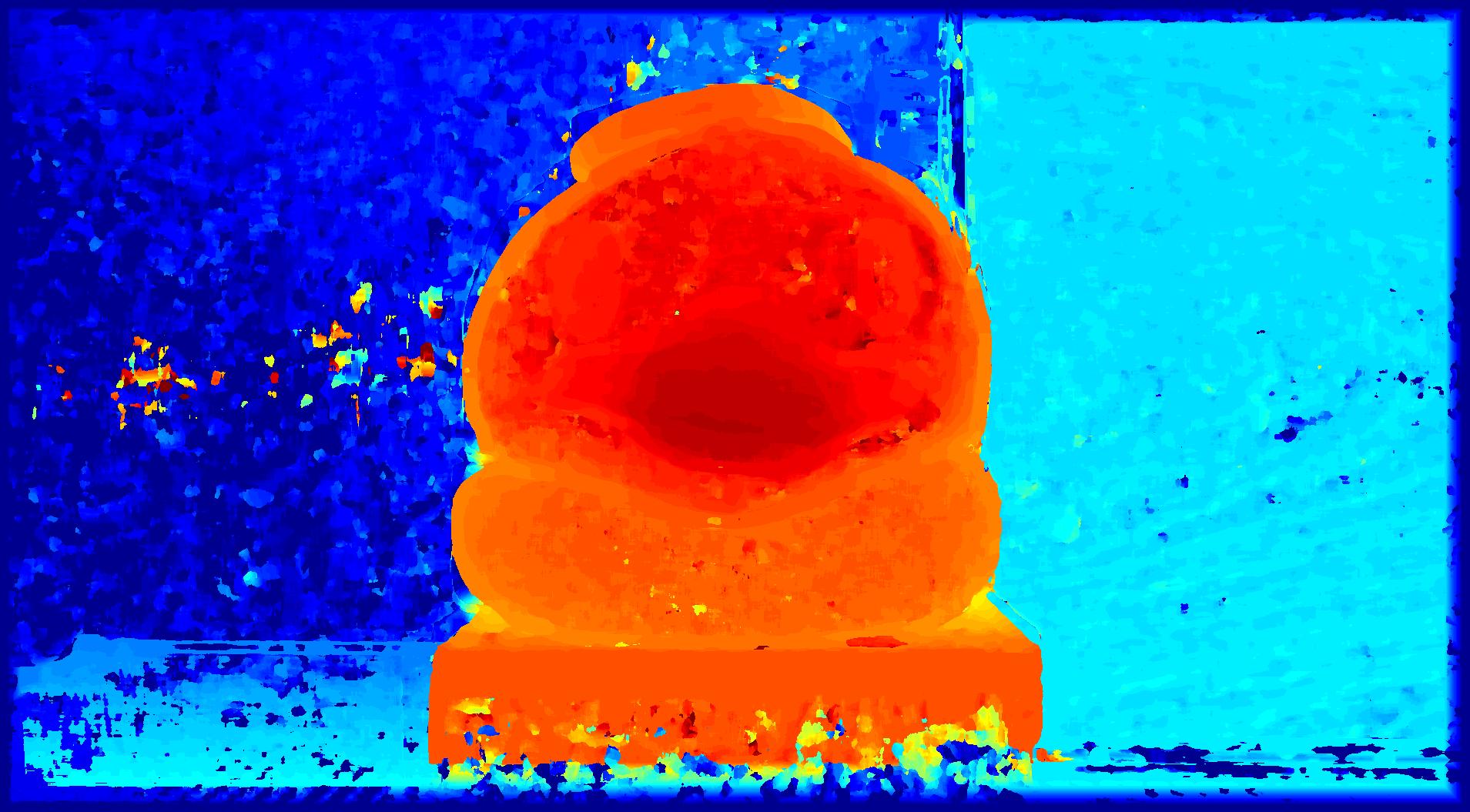}
  \caption{Heuristic fusion}
\end{subfigure}
\begin{subfigure}{0.49\columnwidth}
  \centering
  \includegraphics[width=1\columnwidth, trim={2cm 1cm 4cm 1cm}, clip]{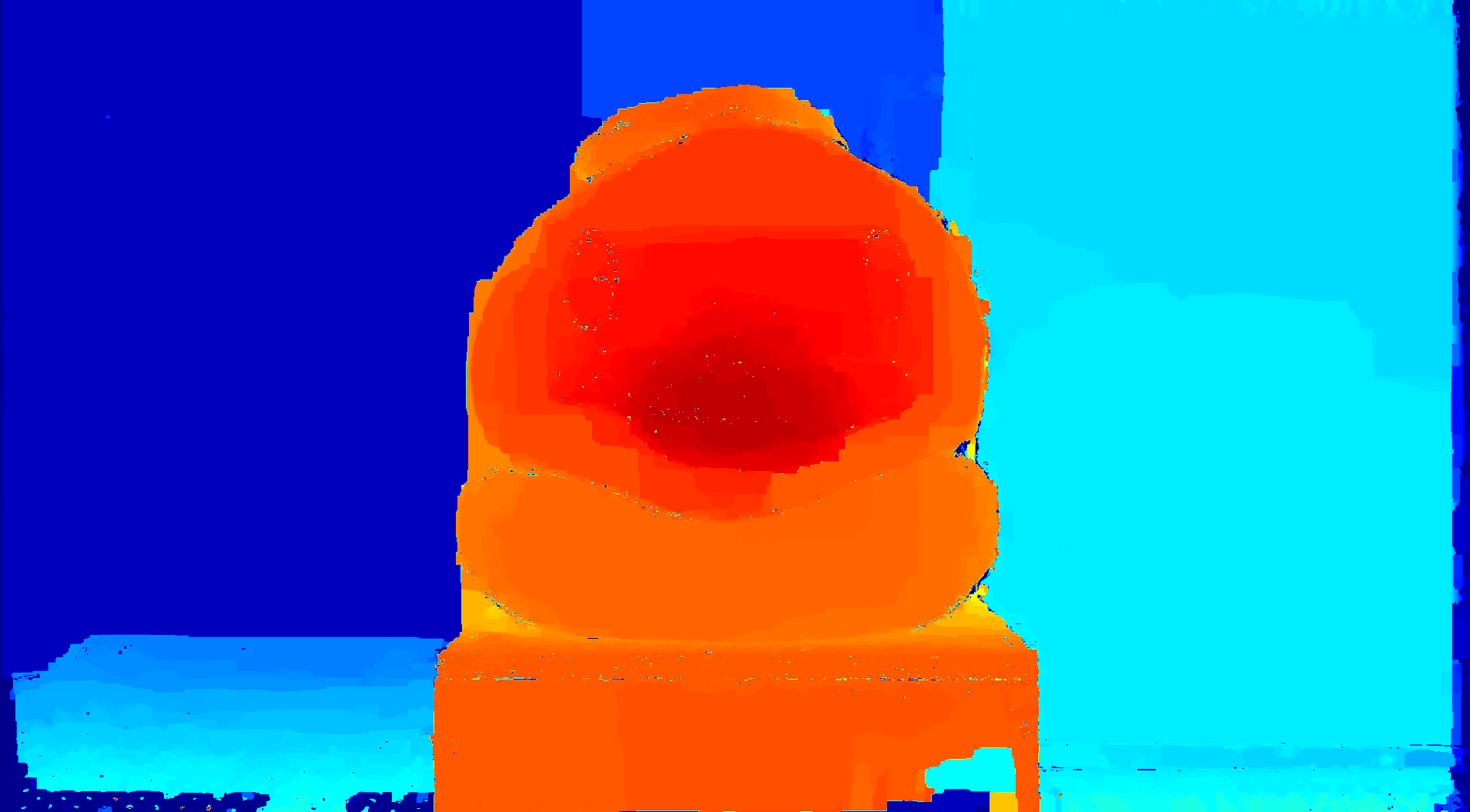}
  \caption{Multiscopic graph cuts}
\end{subfigure}
\caption{The disparity results of various multiscopic algorithms. Block matching with mean, minimum and heuristic SAD cost fusion produce disparity maps (a), (b), (c) respectively and (d) is using multiscopic graph cuts with heuristic fusion. Note that the metal table surface in the bottom left corner and bottom right corner is reconstructed well now.}
\label{fig:multi}
\end{figure}

The visual result of using $c_{\text{ave}}$ shown in Fig.~\ref{fig:multi}(a) indicates that it does remove much noise and reconstruct the reflective tabletop better, but the result is still affected by occlusion areas. For the center image, some regions can not be seen in some surrounding images. For instance, the region to the left of the toy cannot be seen in the right image. Thus the cost $c_{\text{SAD1}}$ for this region would be large and may affect the overall data term $c_{\text{ave}}$. Therefore we consider another fusion strategy by choosing the smallest one when combining the four parts:
\begin{equation}
  c_{\text{min}}(u,v,d)=\min\{c_{\text{SAD1}},c_{\text{SAD2}},c_{\text{SAD3}},c_{\text{SAD4}}\}.
  \label{equ:min}
\end{equation}

The visual result with $c_{\text{min}}$ is presented in Fig.~\ref{fig:multi}(b). We can see that the occlusion region is reconstructed clearly but the noise is persistent in some areas. To overcome this, we design a heuristic fusion strategy. First we sort the four costs on each pixel and use three smallest costs $c^{\text{\Romannum{1}}}, c^{\text{\Romannum{2}}}, c^{\text{\Romannum{3}}}$ ($c^{\text{\Romannum{1}}}$ is the smallest). Then we remove the second largest cost if it is much larger than the other two:
\begin{equation}
c_{\text{heu}}(u,v,d)=
\begin{cases}
\frac{1}{2}(c^{\text{\Romannum{1}}}+c^{\text{\Romannum{2}}}), \quad & \text{if} \ c^{\text{\Romannum{3}}}>c^{\text{\Romannum{2}}} \times 3\\
\frac{1}{3}(c^{\text{\Romannum{1}}}+c^{\text{\Romannum{2}}}+c^{\text{\Romannum{3}}}), \quad & \text{otherwise}
\end{cases}
,
\label{equ:heuristic}
\end{equation}
which leads to a cleaner disparity map as shown in Fig.~\ref{fig:multi}(c).

\subsection{Multiscopic Graph Cuts}

The optimization using graph cuts in multiscopic vision is similar to the two-frame stereo matching except modification on the data term and the smoothness term.

\textbf{Data term} $E_{\text{data}}$ now is also an integration of four parts:
\begin{equation}
\begin{aligned}
c_{\text{BT}1}(u,v,d)=\max\{0, I_c(u,v)-I_r^{\min}(u-d,v), \\ I_r^{\max}(u-d,v)-I_c(u,v)\},\\
c_{\text{BT}2}(u,v,d)=\max\{0, I_c(u,v)-I_l^{\min}(u+d,v), \\ I_l^{\max}(u+d,v)-I_c(u,v)\},\\
c_{\text{BT}3}(u,v,d)=\max\{0, I_c(u,v)-I_t^{\min}(u,v+d), \\ I_t^{\max}(u,v+d)-I_c(u,v)\},\\
c_{\text{BT}4}(u,v,d)=\max\{0, I_c(u,v)-I_b^{\min}(u,v-d), \\ I_t^{\max}(u,v-d)-I_c(u,v)\},
\end{aligned}
\end{equation}
where $I_r^{\min}, I_r^{\max}, I_l^{\min}, I_l^{\max}, I_t^{\min}, I_t^{\max}, I_b^{\min}, I_b^{\max}$ are the smallest and largest values on the subpixel neighborhood in the right, left, top, bottom image respectively. These four costs are then merged using the same heuristic rule to get the fused cost $c_{\text{GC}}(u,v,d)$.

\textbf{Smoothness term} $E_{\text{smoothness}}$ now should take the color and disparity discontinuity in every image into account. If two adjacent pixels $p_1, p_2$ in the center image have different disparity assignments, then the penalty should be
\begin{equation}
  V=
  \begin{cases}
    \lambda_1 \Delta d \ \max\{|I_c(p_1)-I_c(p_2)|, |I_r(q_1)-I_r(q_2)|,\\
    \qquad \qquad \ |I_l(m_1)-I_l(m_2)|,
    |I_t(n_1)-I_t(n_2)|,\\
    \qquad \qquad \ |I_b(k_1)-I_b(k_2)|
    \}<\theta \\
    \lambda_2 \Delta d \quad \text{otherwise}
  \end{cases},
\end{equation}
where $q_1, q_2, m_1, m_2, n_1, n_2, k_1, k_2$ are the corresponding pixels of $p_1,p_2$ in the right, left, top, bottom images respectively.

With the new energy, the result of the multiscopic graph cuts with the same hyper-parameters is displayed in Fig.~\ref{fig:multi}(d). Compared with the stereo graph cuts displayed in Fig.~\ref{fig:stereo}(b), the occlusion parts and reflective tabletop are reconstructed much better and the noise is better suppressed.


\section{EXPERIMENTS}
\label{sec:exp}

In this section, we present the details of our system setup and experiments. The quantitative evaluation on the Middlebury Stereo Dataset and the qualitative test on real robot experiments are demonstrated to compare the multiscopic vision with the two-frame stereo matching.

\begin{figure}[h]
\centering
  \includegraphics[width=0.6\columnwidth, trim={0cm 0cm 0cm 0cm}, clip]{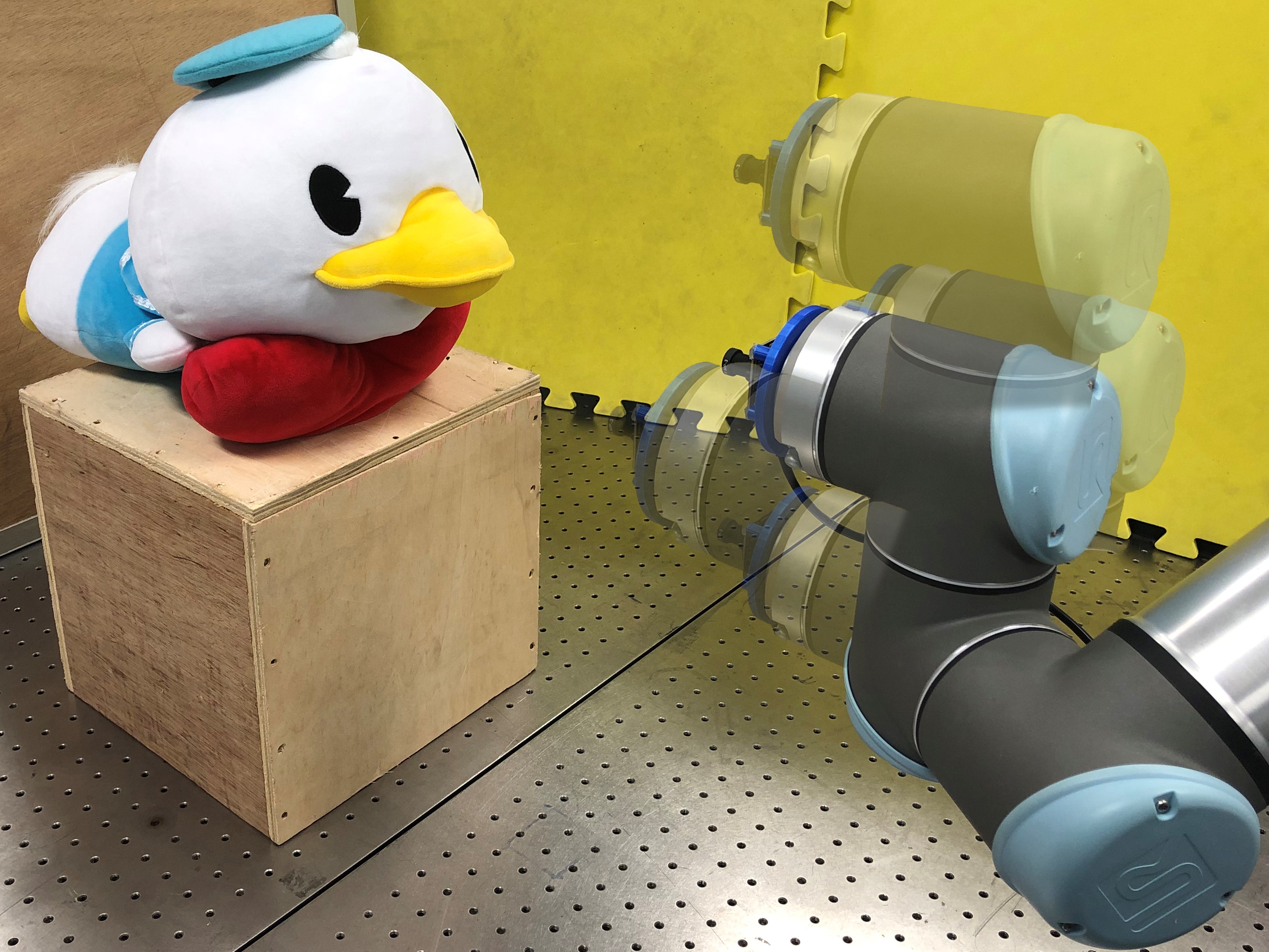}
\caption{The camera is mounted on the end of a robot arm and moved horizontally and vertically to take pictures from different views.}
\label{fig:robot}
\end{figure}

\begin{figure*}[t]
\centering
\begin{subfigure}{0.33\columnwidth}
  \centering
  \includegraphics[width=1.02\columnwidth, trim={0cm 0cm 0cm 0cm}, clip]{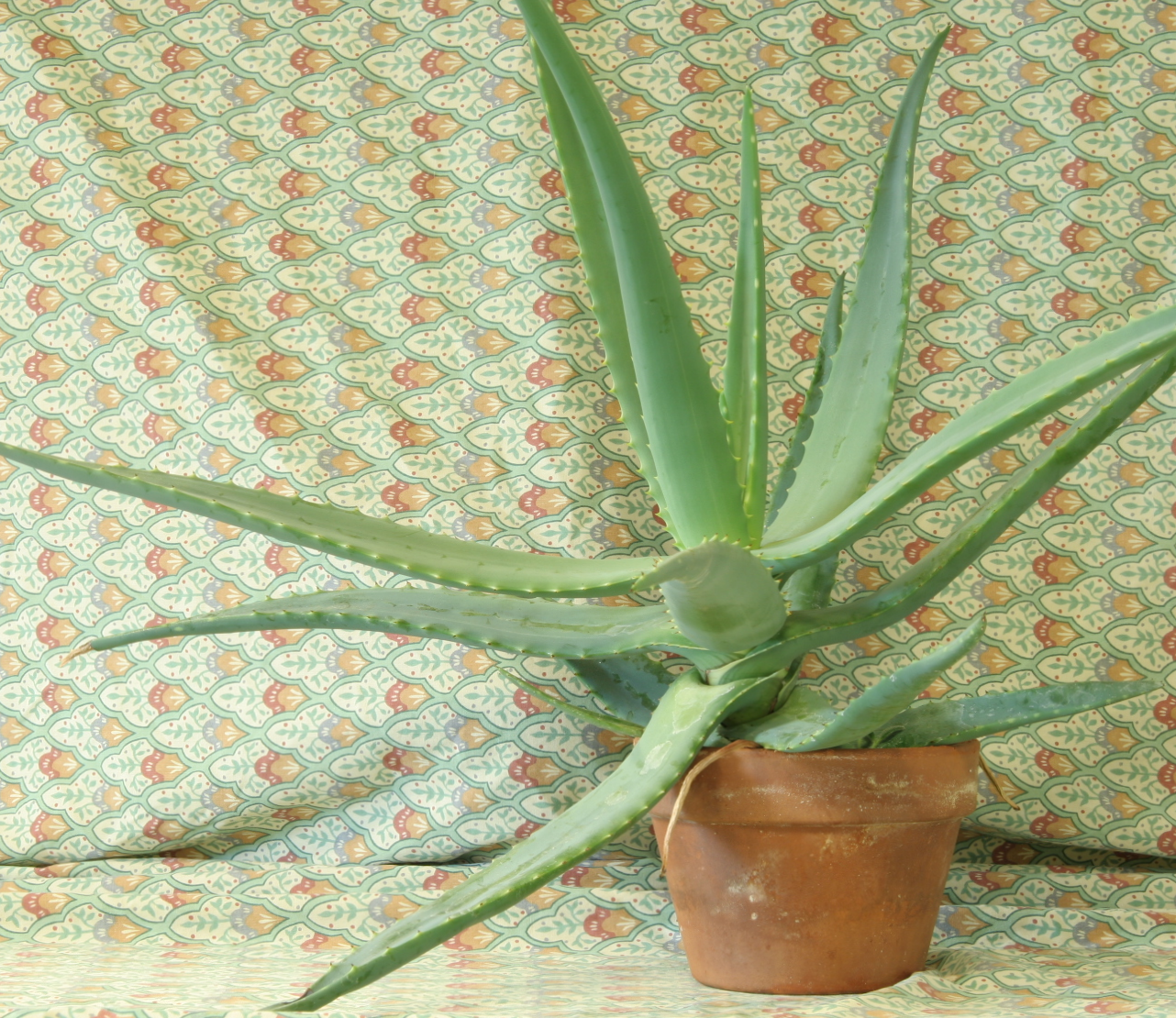}
  \caption{Reference image}
\end{subfigure}
\begin{subfigure}{0.33\columnwidth}
  \centering
  \includegraphics[width=1.02\columnwidth, trim={0cm 0cm 0cm 0cm}, clip]{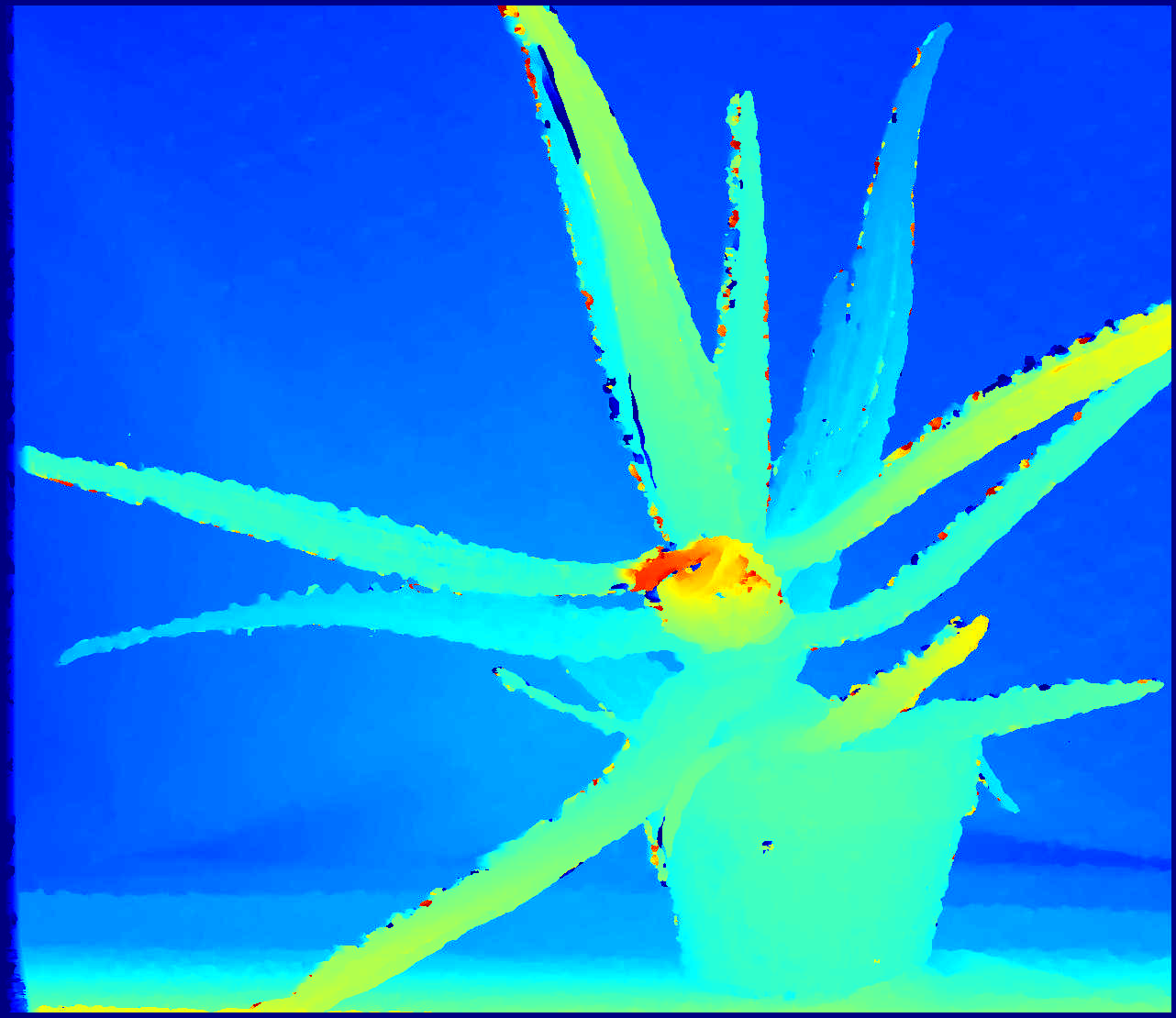}
  \caption{Stereo BM}
\end{subfigure}
\begin{subfigure}{0.33\columnwidth}
  \centering
  \includegraphics[width=1.02\columnwidth, trim={0cm 0cm 0cm 0cm}, clip]{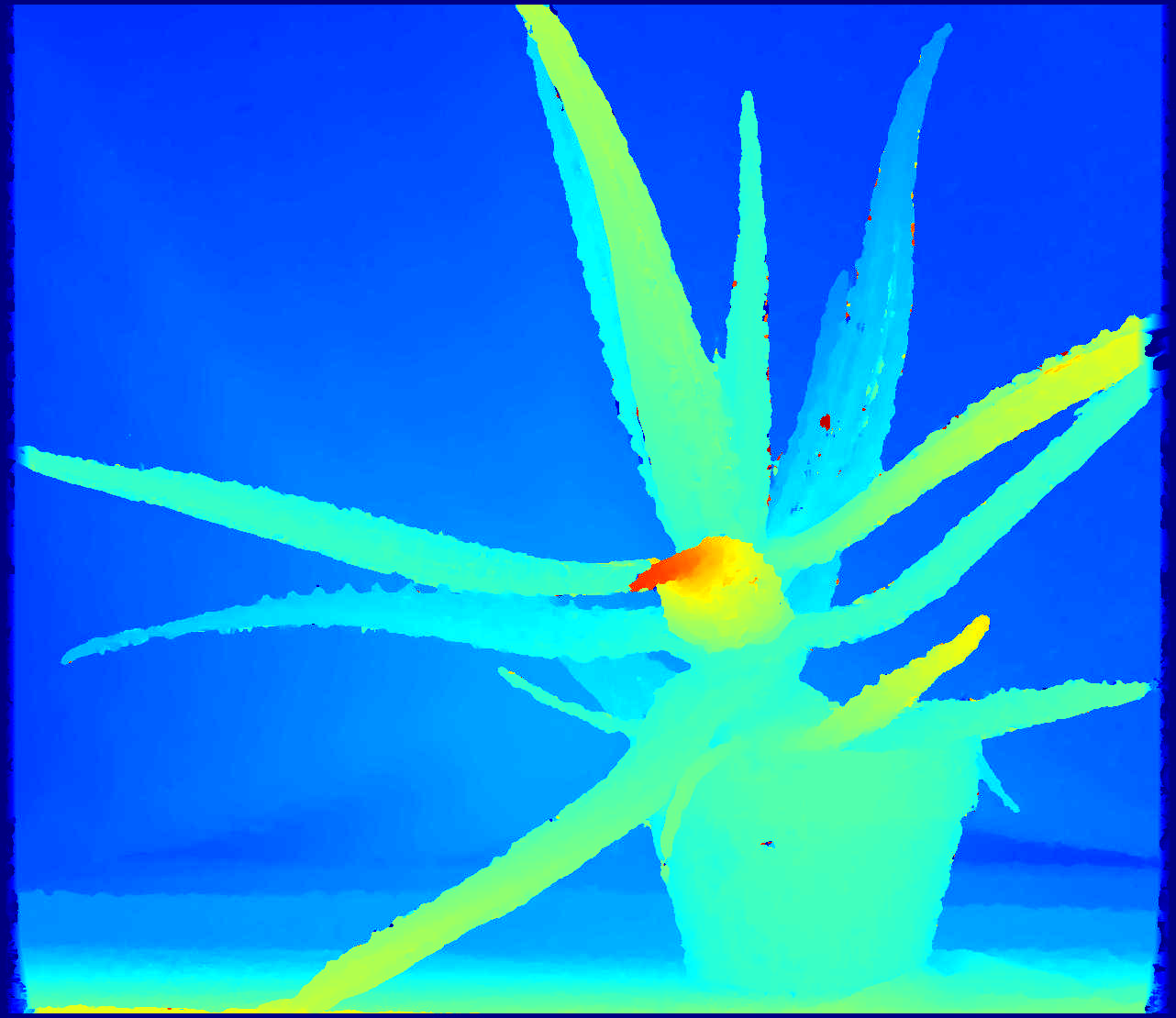}
  \caption{Multiscopic BM}
\end{subfigure}
\begin{subfigure}{0.33\columnwidth}
  \centering
  \includegraphics[width=1.02\columnwidth, trim={0cm 0cm 0cm 0cm}, clip]{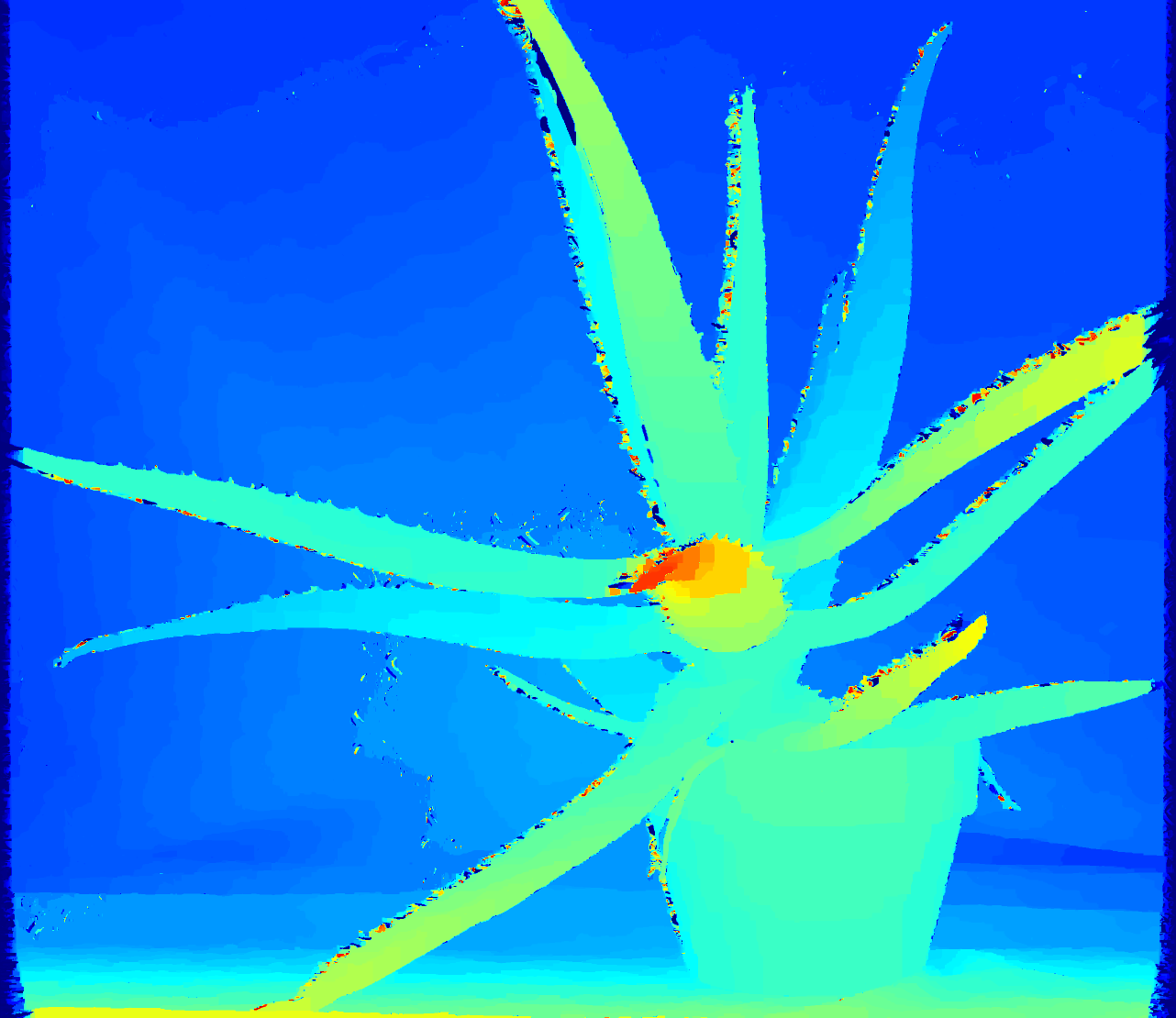}
  \caption{Stereo GC}
\end{subfigure}
\begin{subfigure}{0.33\columnwidth}
  \centering
  \includegraphics[width=1.02\columnwidth, trim={0cm 0cm 0cm 0cm}, clip]{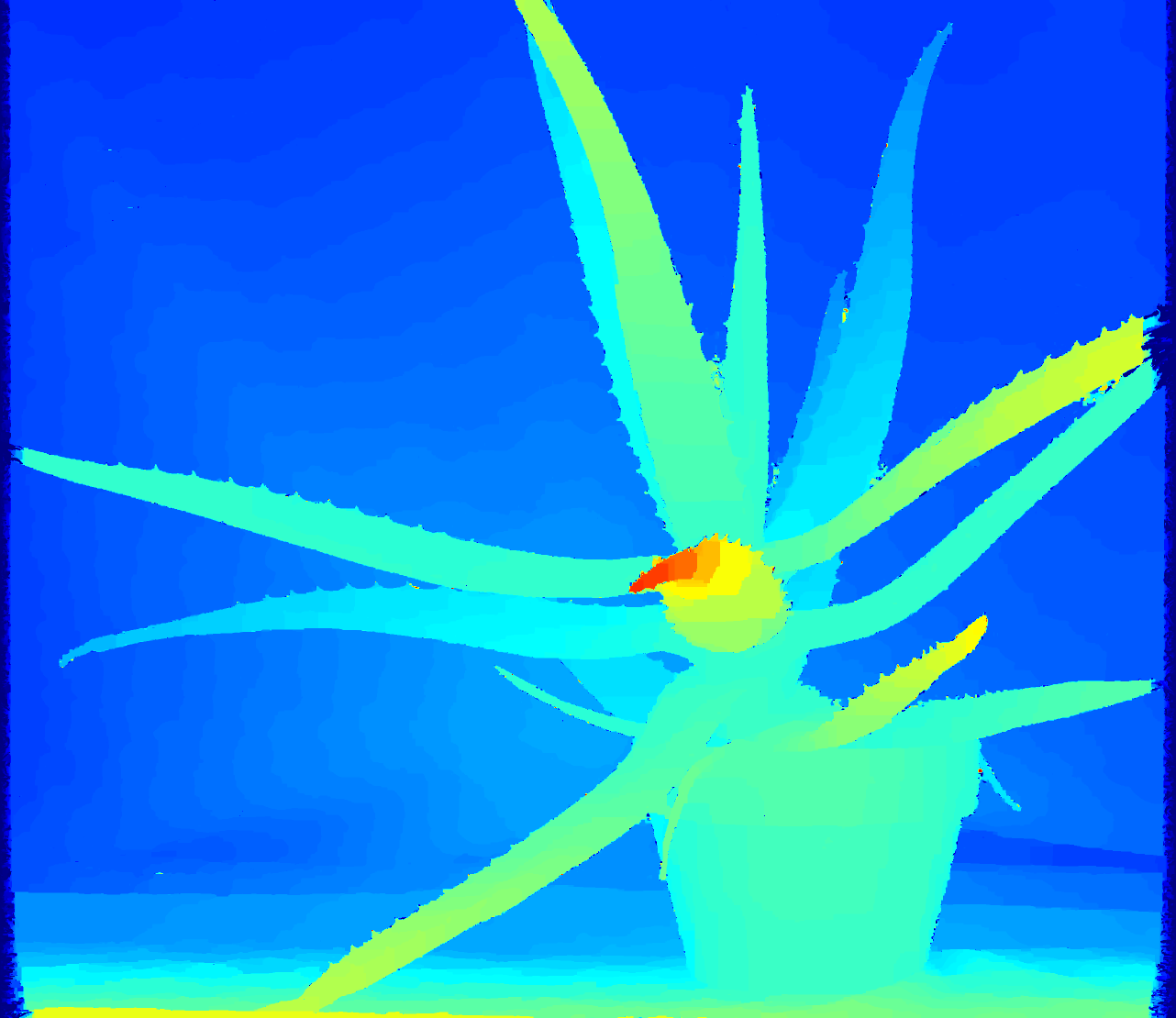}
  \caption{Multiscopic GC}
\end{subfigure}
\begin{subfigure}{0.33\columnwidth}
  \centering
  \includegraphics[width=1.02\columnwidth, trim={0cm 0cm 0cm 0cm}, clip]{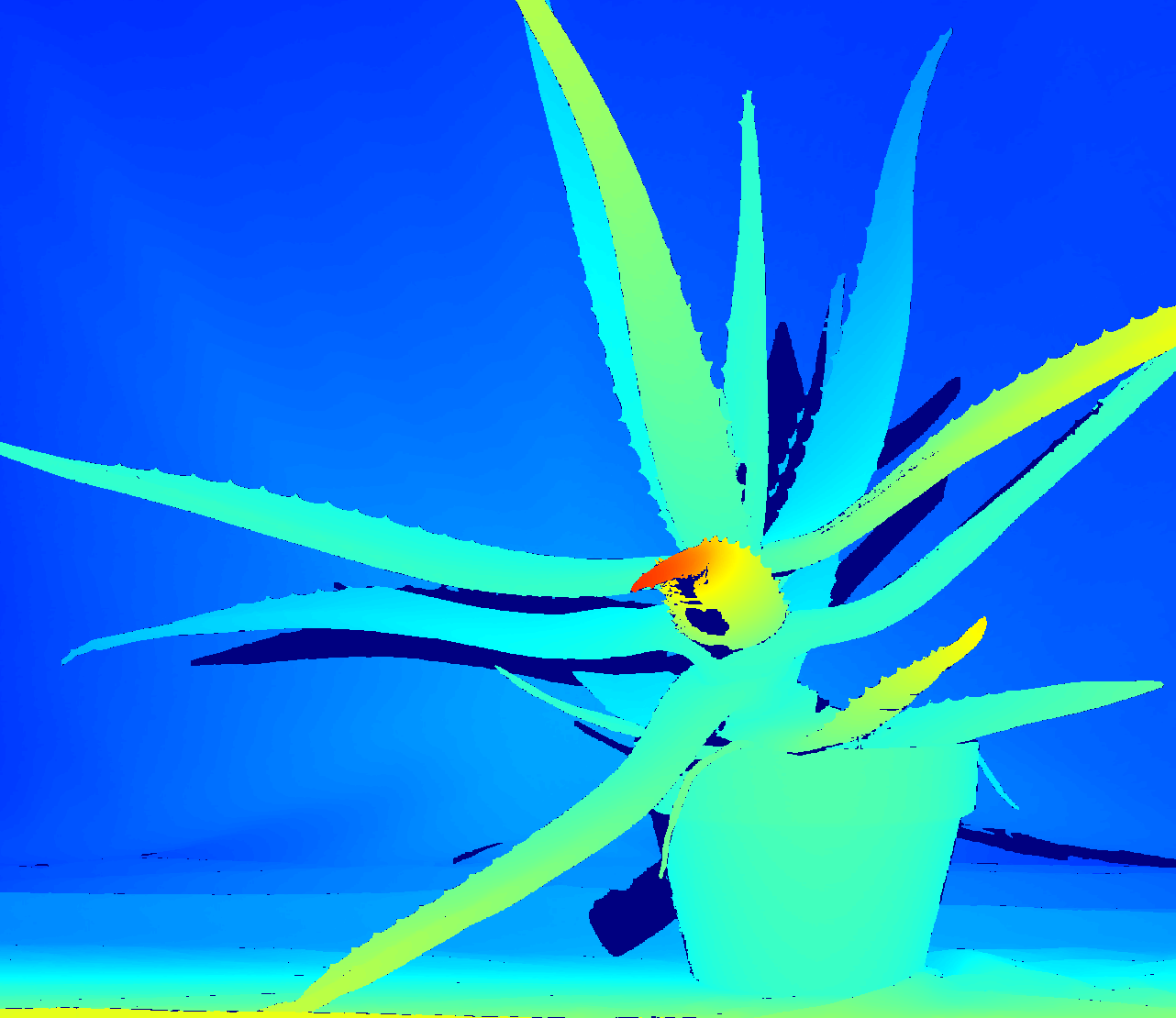}
  \caption{Ground truth}
\end{subfigure}

\begin{subfigure}{0.33\columnwidth}
  \centering
  \includegraphics[width=1.02\columnwidth, trim={0cm 0cm 0cm 0cm}, clip]{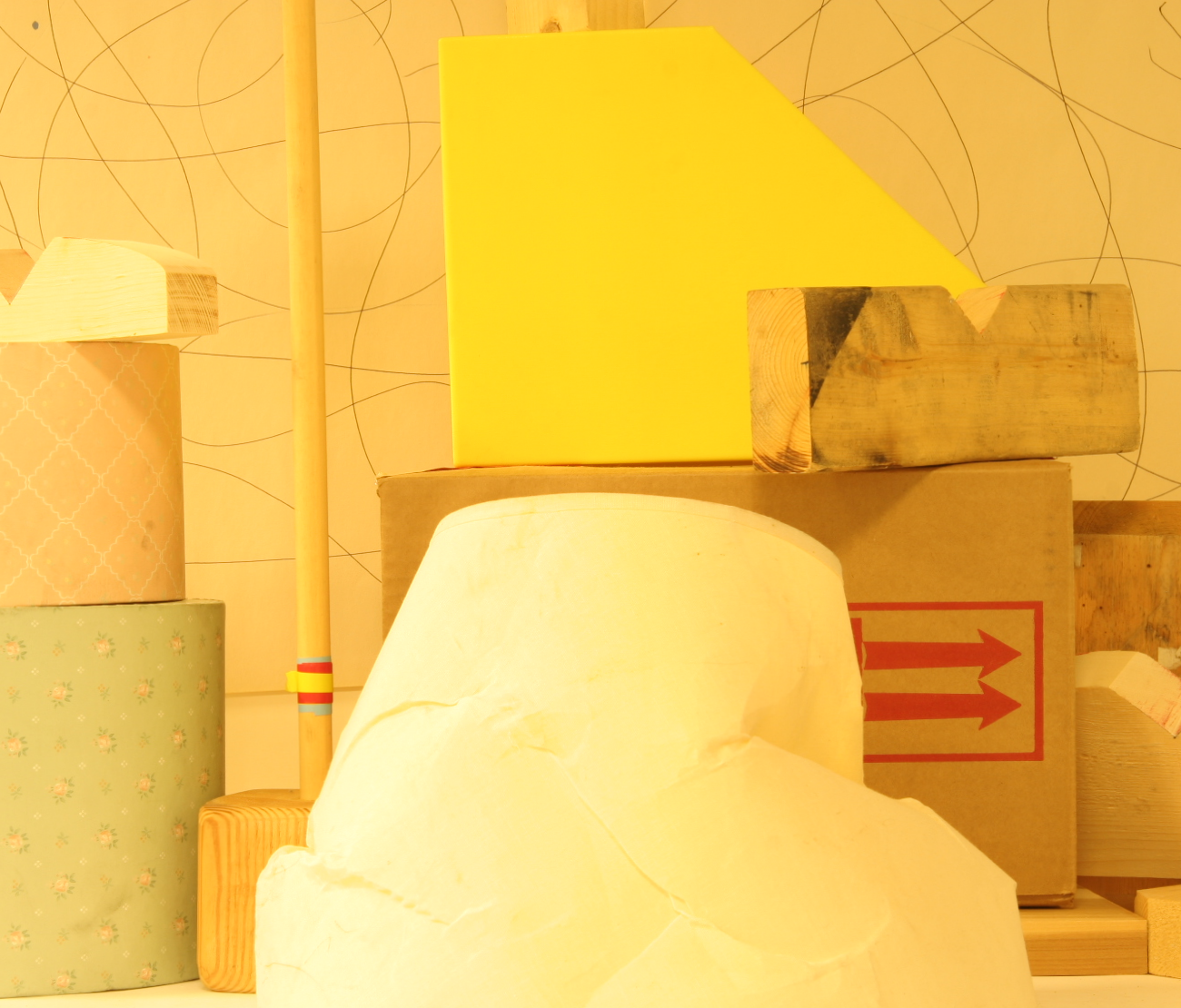}
  \caption{Reference image}
\end{subfigure}
\begin{subfigure}{0.33\columnwidth}
  \centering
  \includegraphics[width=1.02\columnwidth, trim={0cm 0cm 0cm 0cm}, clip]{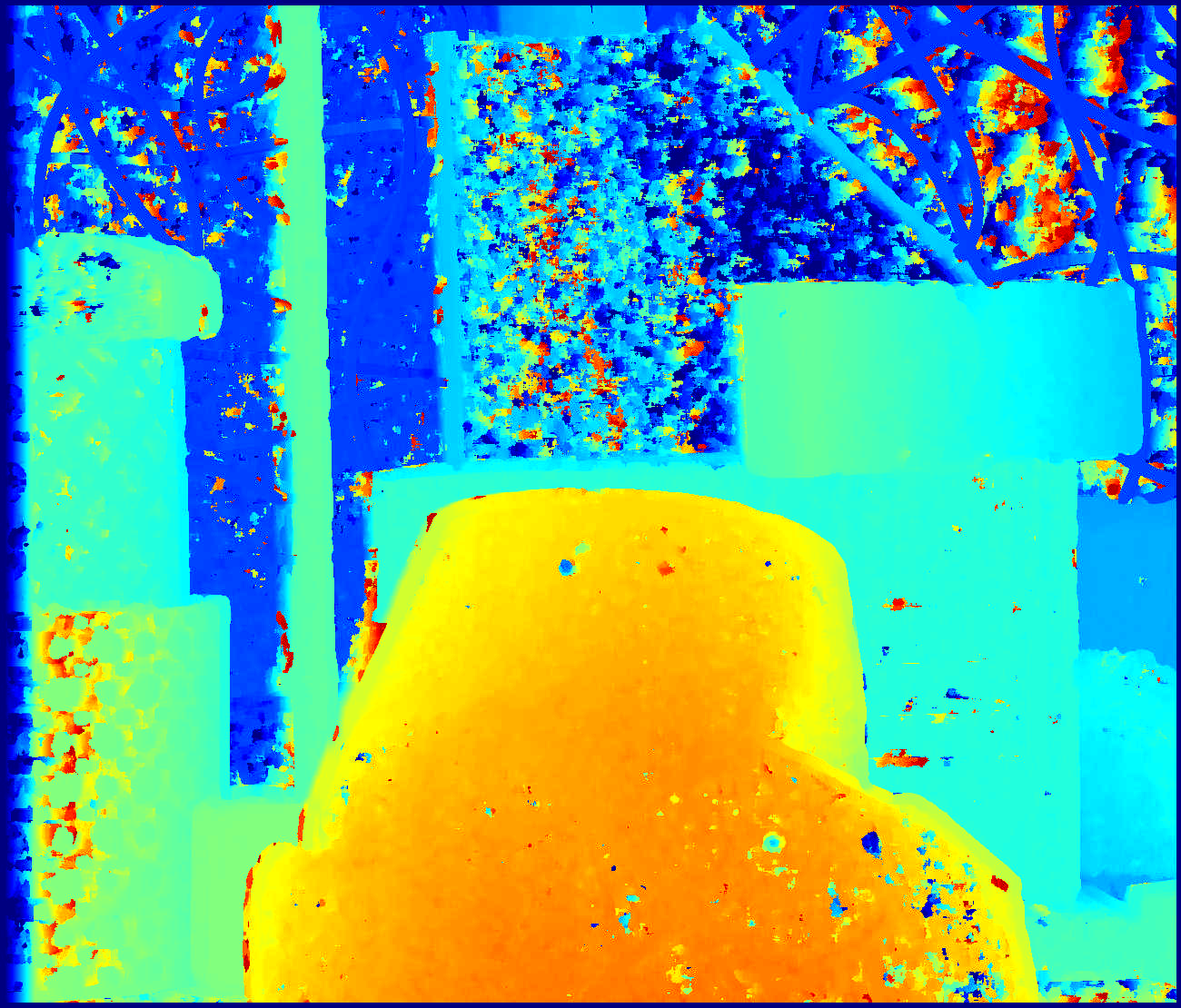}
  \caption{Stereo BM}
\end{subfigure}
\begin{subfigure}{0.33\columnwidth}
  \centering
  \includegraphics[width=1.02\columnwidth, trim={0cm 0cm 0cm 0cm}, clip]{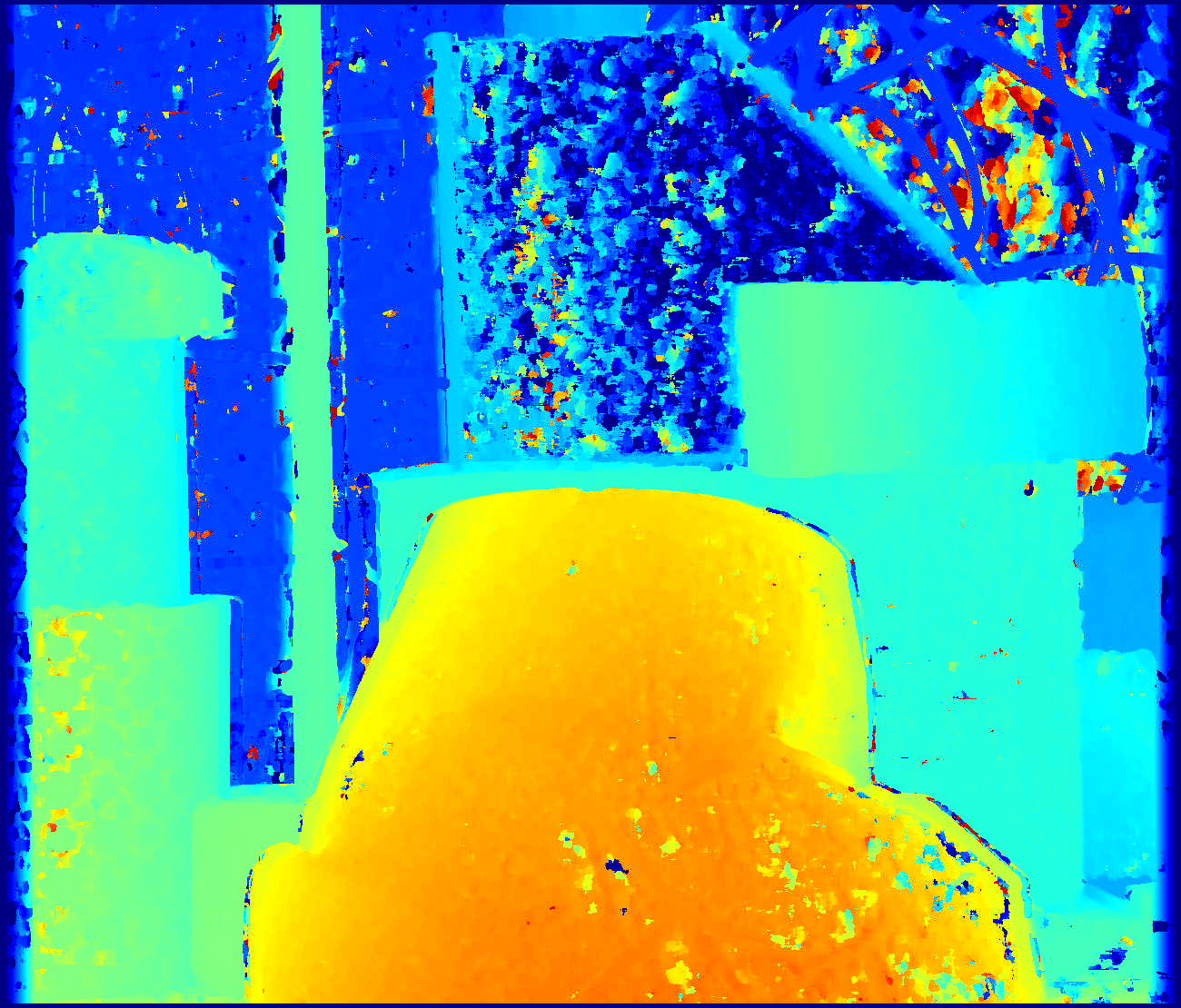}
  \caption{Multiscopic BM}
\end{subfigure}
\begin{subfigure}{0.33\columnwidth}
  \centering
  \includegraphics[width=1.02\columnwidth, trim={0cm 0cm 0cm 0cm}, clip]{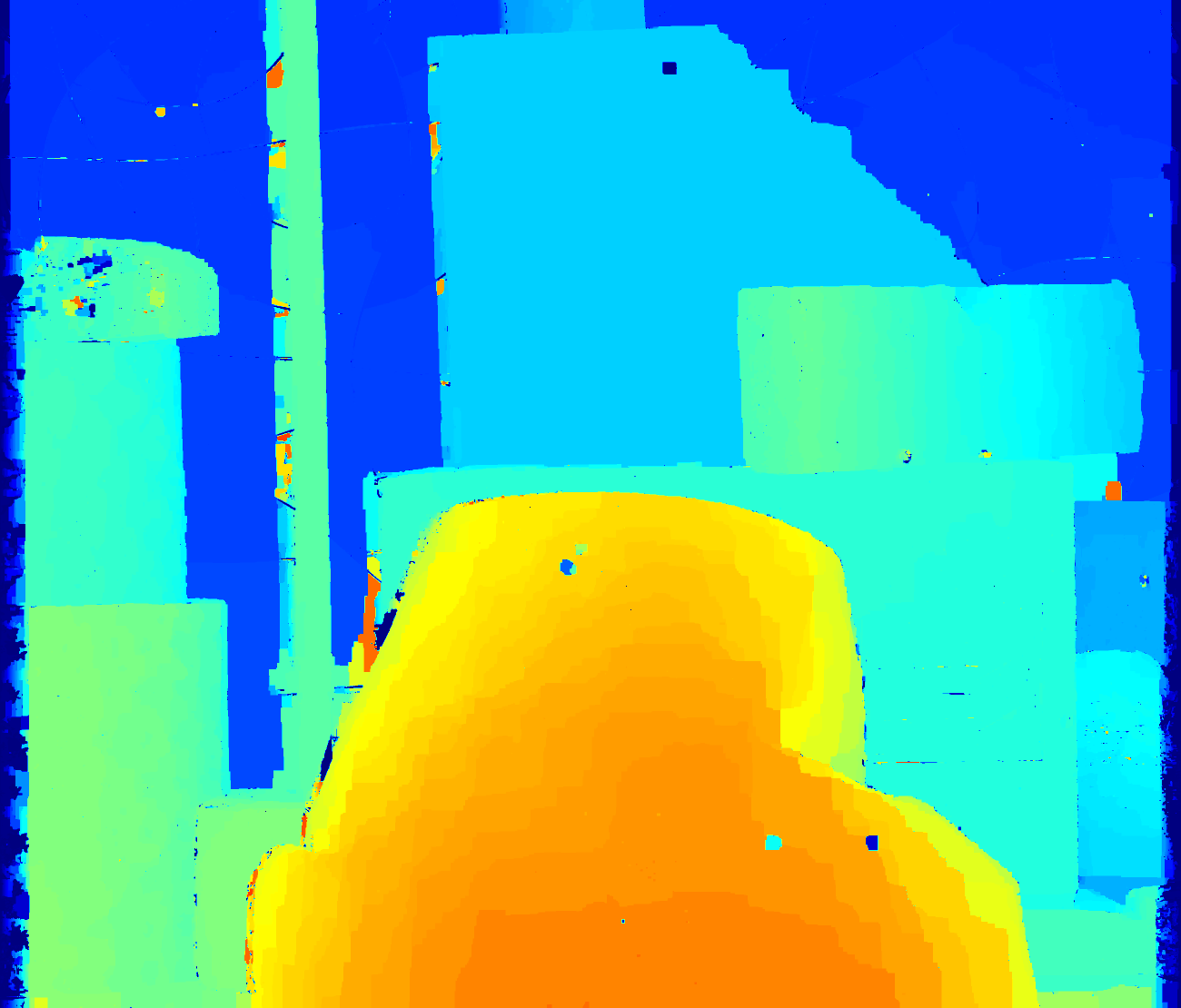}
  \caption{Stereo GC}
\end{subfigure}
\begin{subfigure}{0.33\columnwidth}
  \centering
  \includegraphics[width=1.02\columnwidth, trim={0cm 0cm 0cm 0cm}, clip]{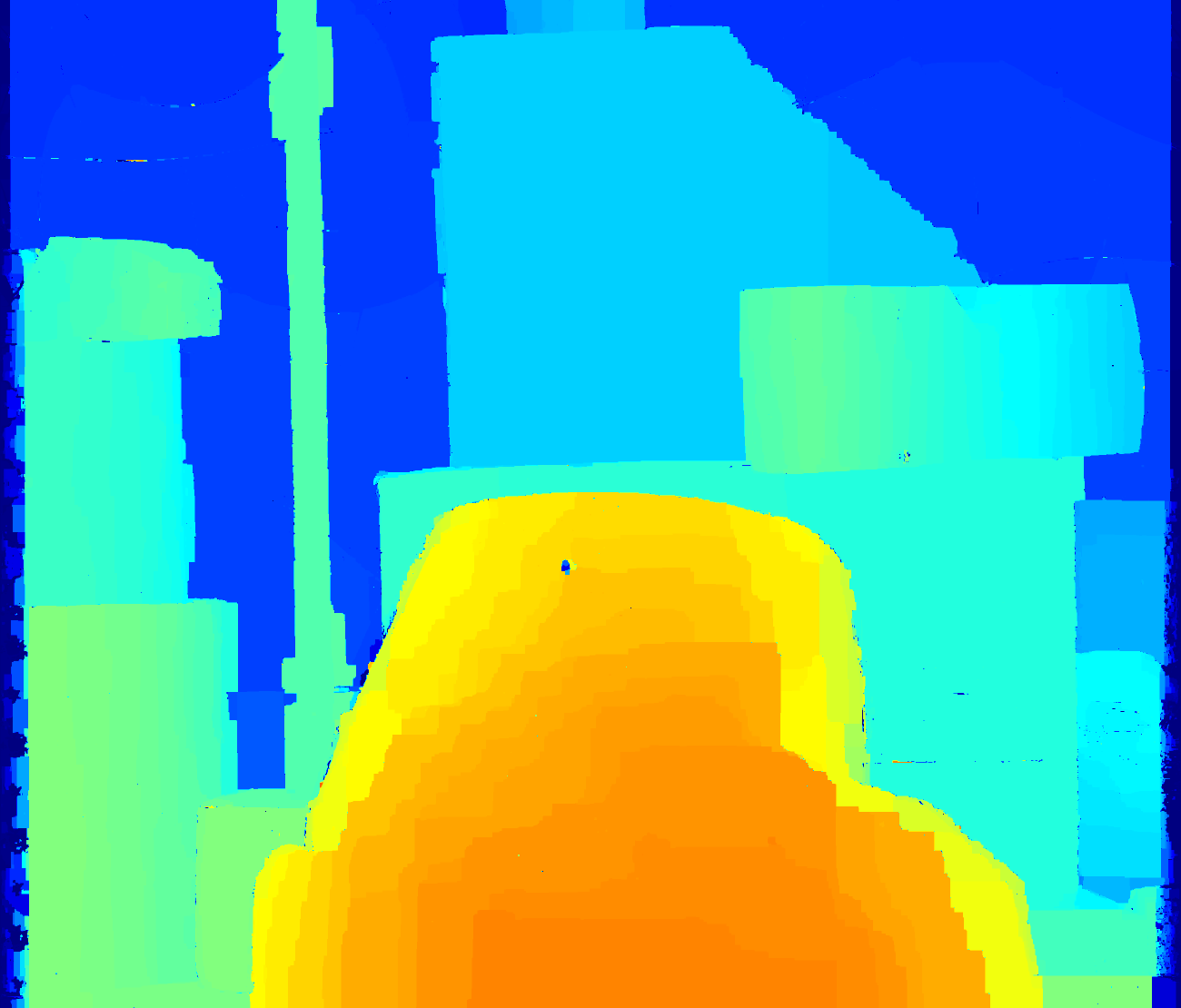}
  \caption{Multiscopic GC}
\end{subfigure}
\begin{subfigure}{0.33\columnwidth}
  \centering
  \includegraphics[width=1.02\columnwidth, trim={0cm 0cm 0cm 0cm}, clip]{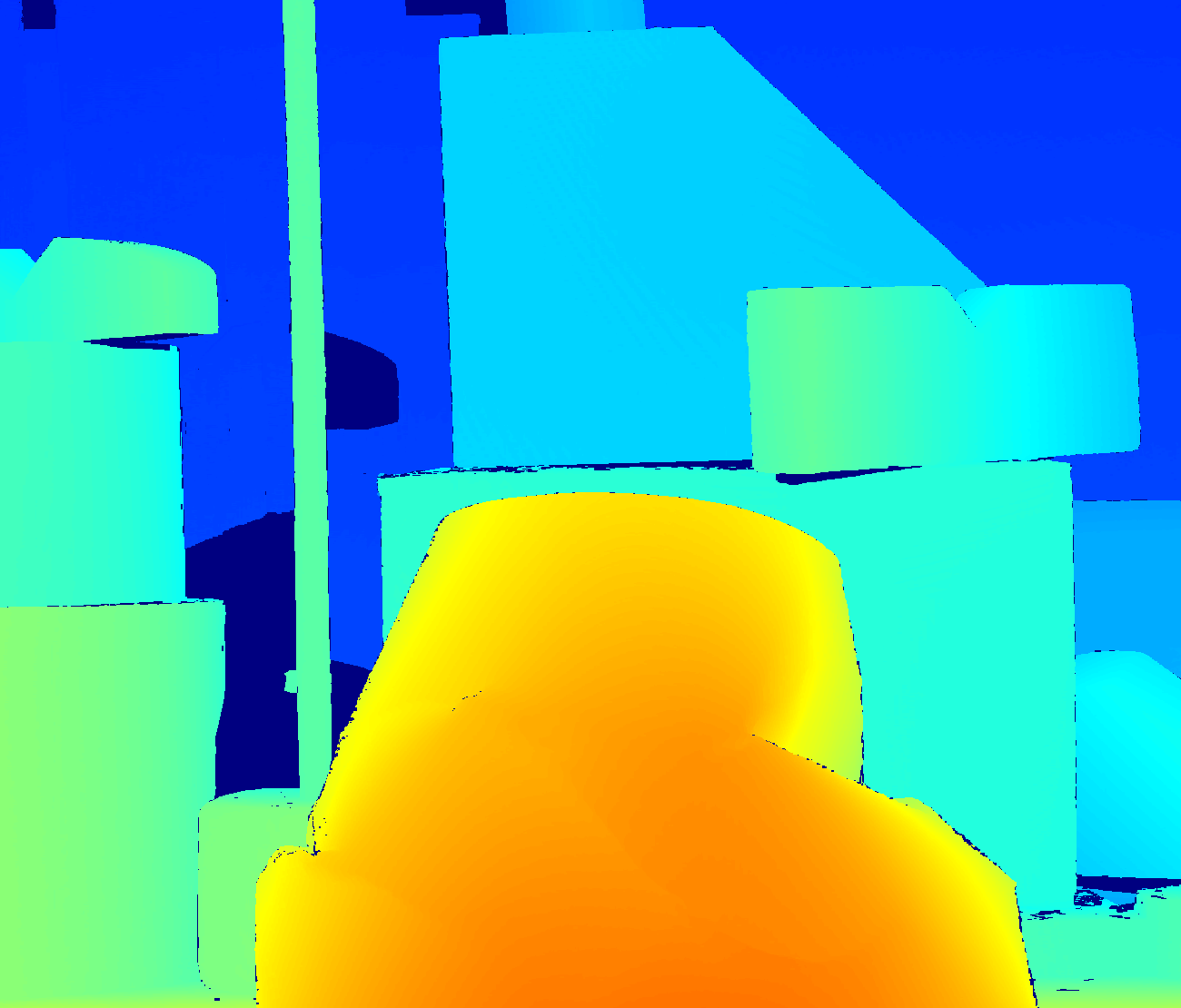}
  \caption{Ground truth}
\end{subfigure}

\caption{The disparity estimation results of different algorithms for two sets of images, Aloe and Lampshade from Middlebury Stereo Datasets. The first image is the reference RGB image, i.e., the left image for stereo algorithms and the center image for multiscopic algorithms. Two images are used for stereo algorithms and three images are used for multiscopic algorithms. BM denotes block matching and GC denotes graph cuts.}
\label{fig:exp_3frame}
\end{figure*}

\subsection{System Setup}

To build the multiscopic vision system with active perception, we mount a monocular camera on the end of a robot arm, as displayed in Fig.~\ref{fig:robot}. The sensor we use is an ordinary USB video camera with Sony IMX322 inside, whose resolution is $1920\times1080$. The robot arm is UR10, a collaborative industrial robot whose repeatability is $\pm0.1\ \text{mm}$. UR10 has six rotating joints, so the end has 6 degrees of freedom. Thus the camera can move freely with any pose.

\begin{figure*}[]
\centering
\begin{subfigure}{0.4\columnwidth}
  \centering
  \includegraphics[width=1\columnwidth, trim={23cm 5cm 23cm 13cm}, clip]{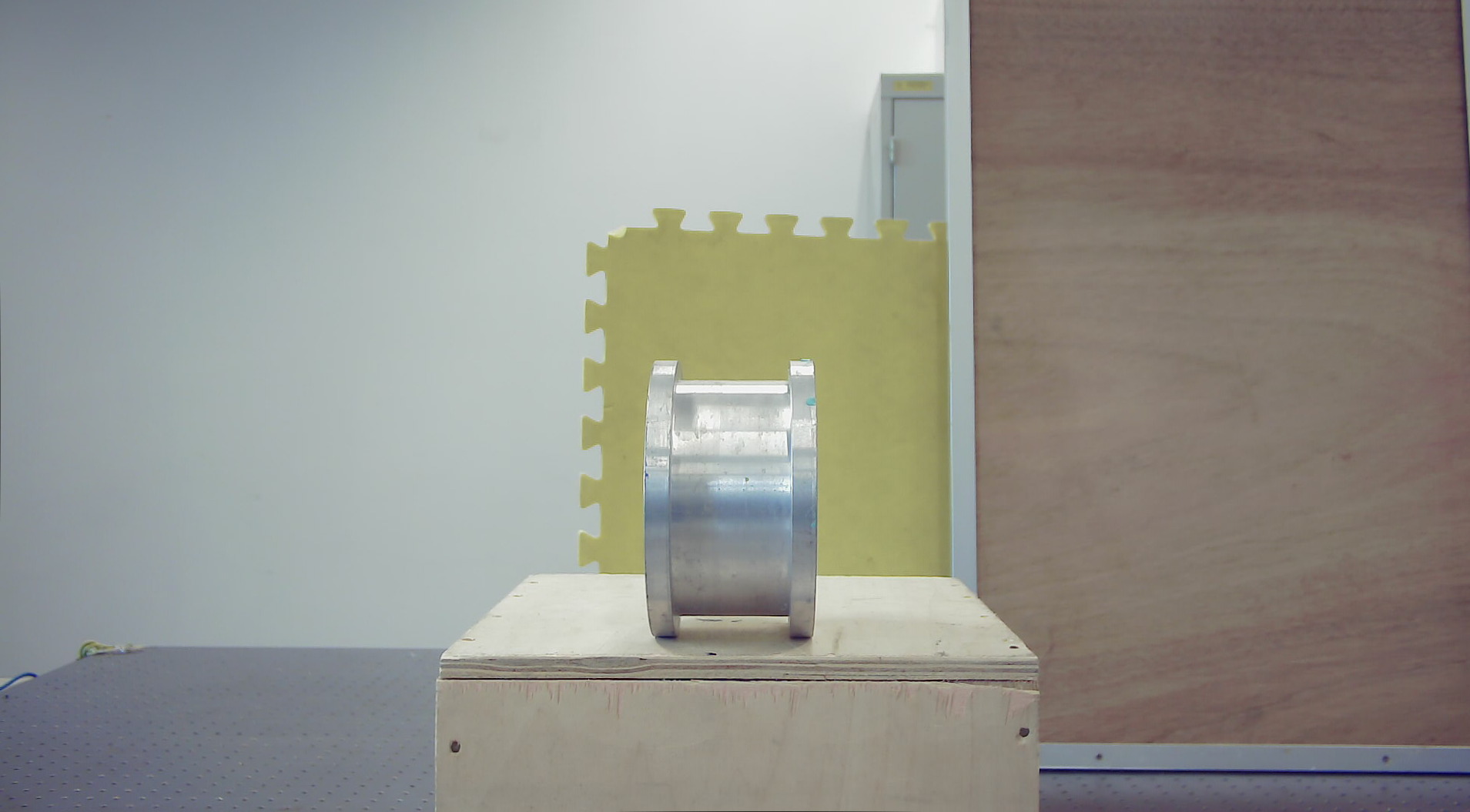}
  \caption{Center image}
\end{subfigure}
\centering
\begin{subfigure}{0.4\columnwidth}
  \centering
  \includegraphics[width=1\columnwidth, trim={23cm 5cm 23cm 13cm}, clip]{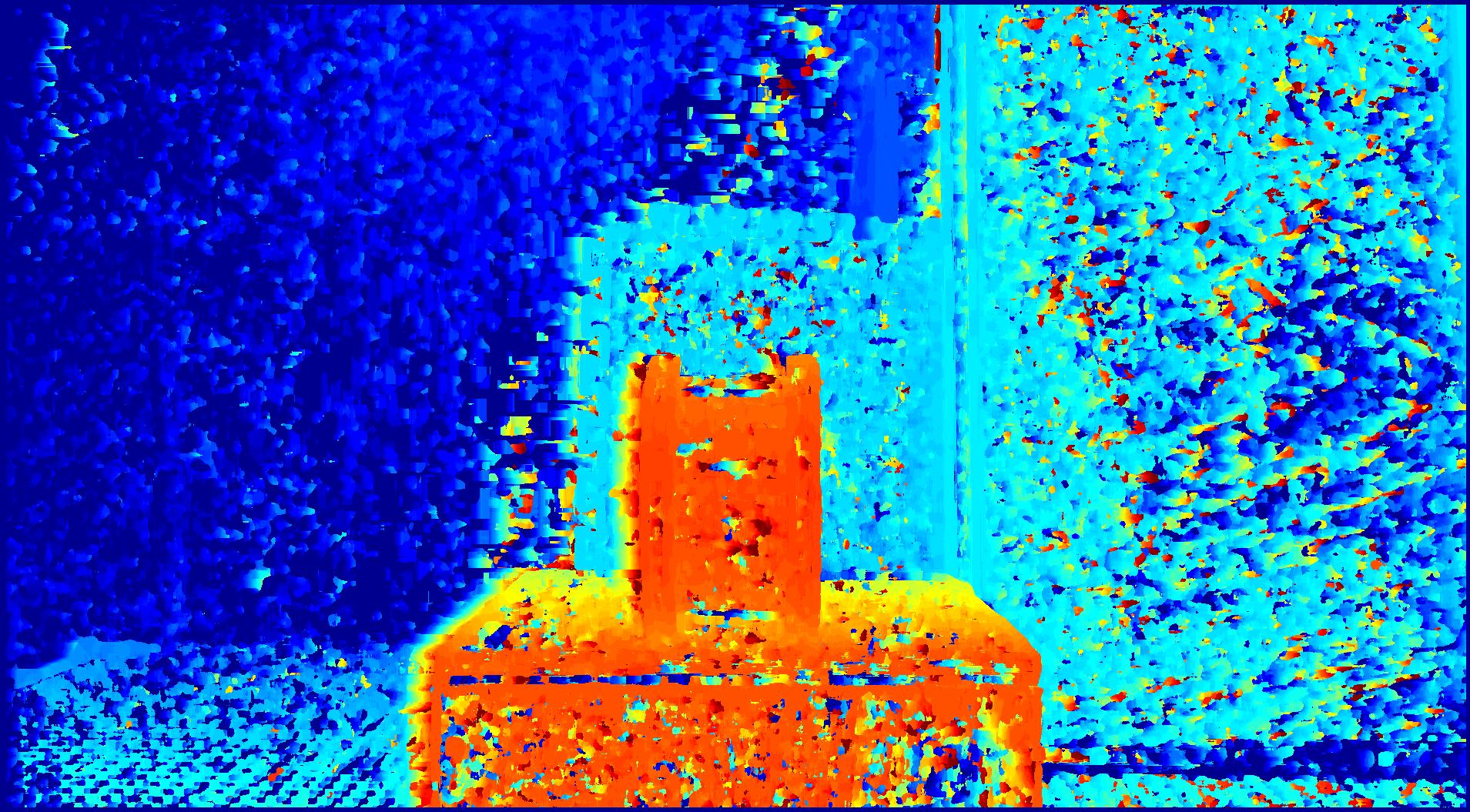}
  \caption{Stereo BM}
\end{subfigure}
\centering
\begin{subfigure}{0.4\columnwidth}
  \centering
  \includegraphics[width=1\columnwidth, trim={23cm 5cm 23cm 13cm}, clip]{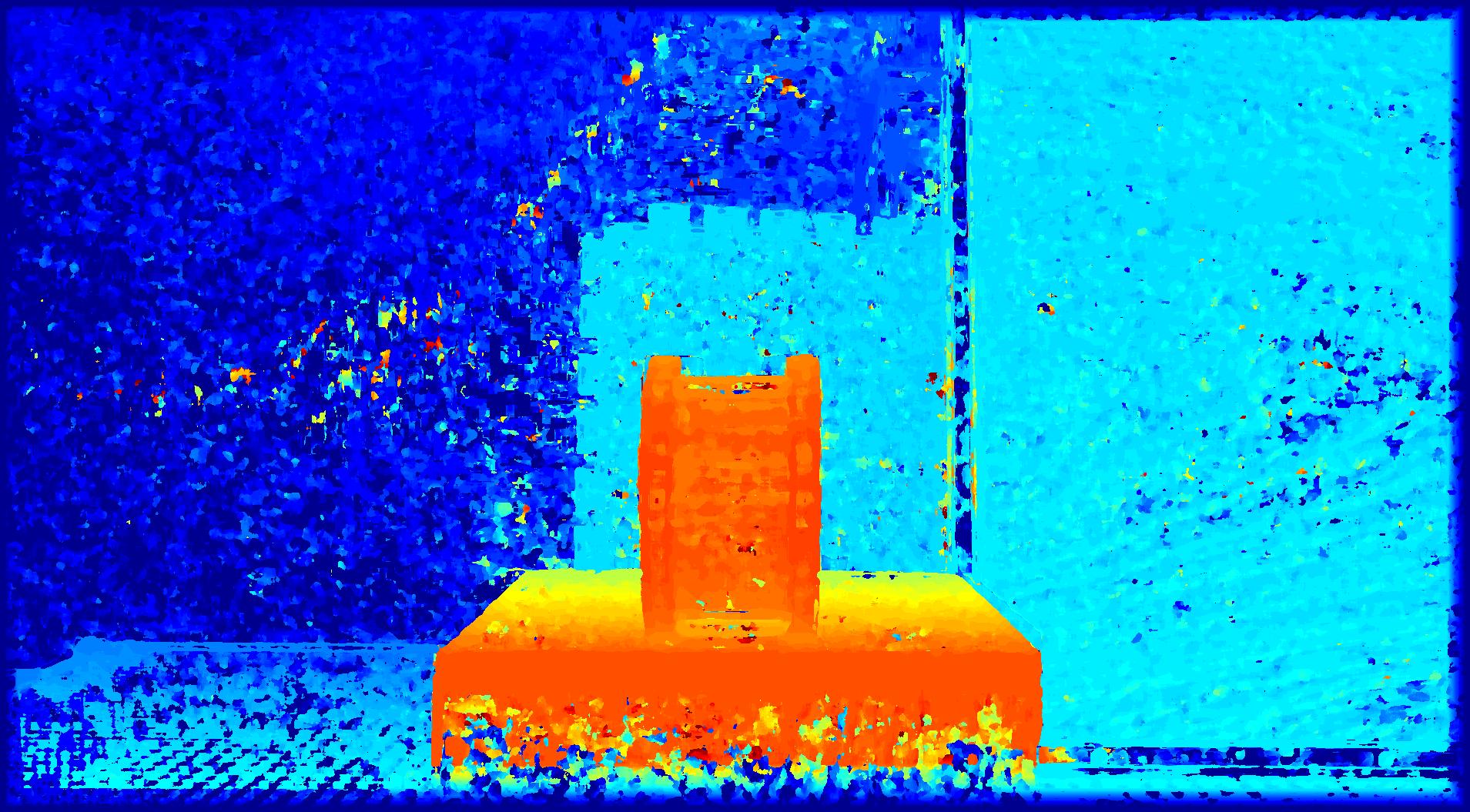}
  \caption{Multiscopic BM}
\end{subfigure}
\begin{subfigure}{0.4\columnwidth}
  \centering
  \includegraphics[width=1\columnwidth, trim={23cm 5cm 23cm 13cm}, clip]{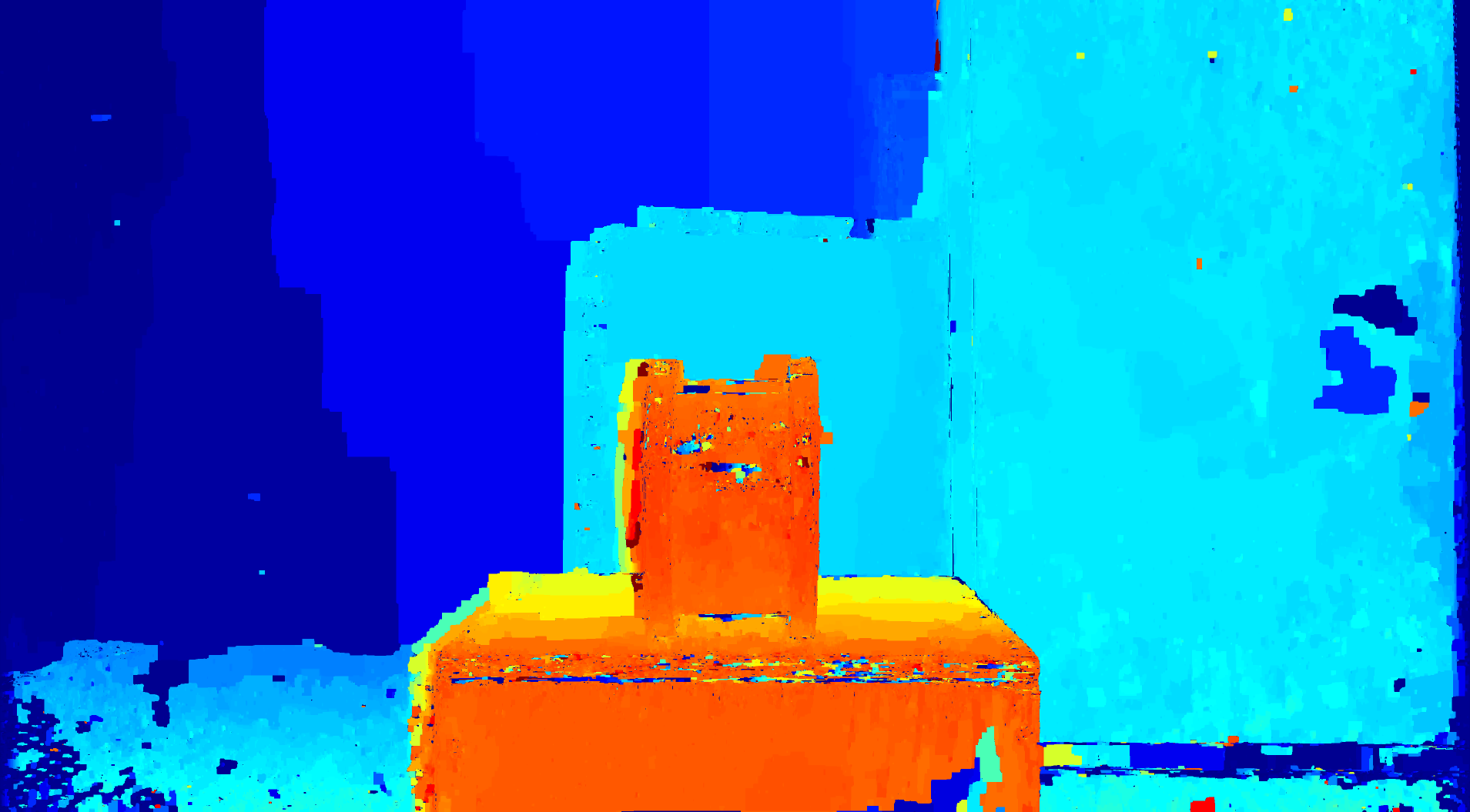}
  \caption{Stereo GC}
\end{subfigure}
\begin{subfigure}{0.4\columnwidth}
  \centering
  \includegraphics[width=1\columnwidth, trim={23cm 5cm 23cm 13cm}, clip]{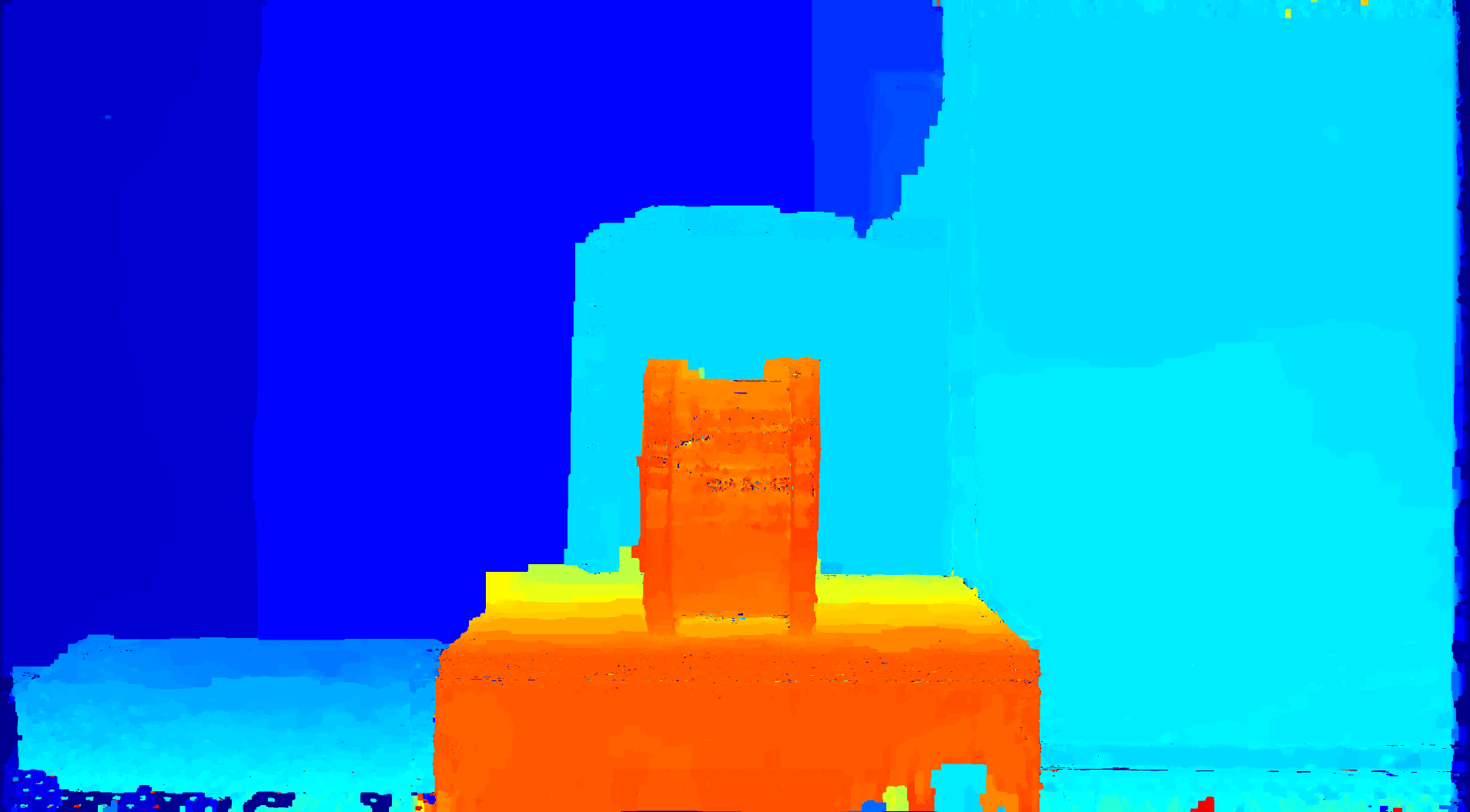}
  \caption{Multiscopic GC}
\end{subfigure}
\caption{The disparity estimation results of different algorithms for a reflective workpiece (zoom in). The multiscopic algorithms use 5 images to do the correspondence searching.}
\label{fig:exp_5frame}
\end{figure*}

To capture a series of images with multiscopic structure, we command the UR10 to move the camera in its image plane, generating a series of co-planar images. For every movement with the same distance, we take one picture of the environment. Thus we can take as many images, and each of these images has the same parallax with its adjacent images. For example, we can take 9 images with 3 rows and 3 columns, which forms a multiscopic array. Also, we can adjust the baseline according to the need. For the sake of simplicity, we use five images in the real robot experiments to estimate disparity, as is demonstrated in Fig.~\ref{fig:exp_5frame}.

To evaluate the performance of our multiscopic vision system, we conduct quantitative evaluation on the Middlebury Stereo Dataset 2006 \cite{hirschmuller2007evaluation} that contains calibrated and rectificated image sequence for depth estimation. We use three adjacent images to compute the disparity in our multiscopic vision system. Note that we can use our active perception system to capture more images and do multiscopic matching with five or even more images.

\begin{figure}[tbh]
\centering
\begin{subfigure}{0.8\columnwidth}
  \centering
  \includegraphics[width=1\columnwidth, trim={0cm 0cm 0cm 0cm}, clip]{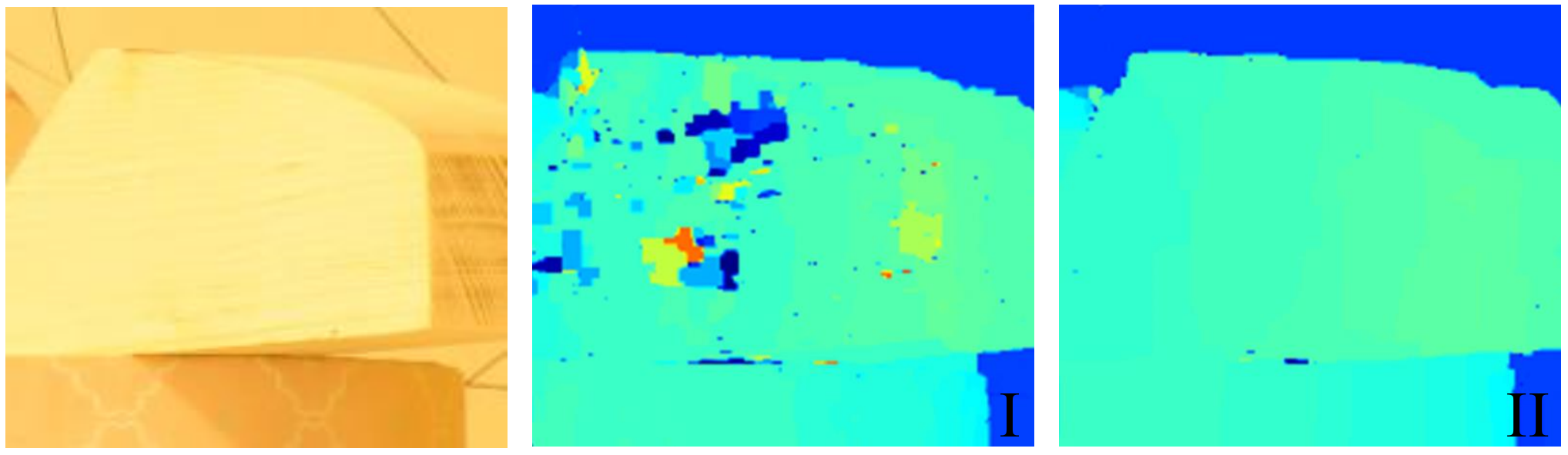}
  \vspace{-0.5cm}
  \caption{Noise}
\end{subfigure}
\begin{subfigure}{0.8\columnwidth}
  \centering
  \includegraphics[width=1\columnwidth, trim={0cm 0cm 0cm 0cm}, clip]{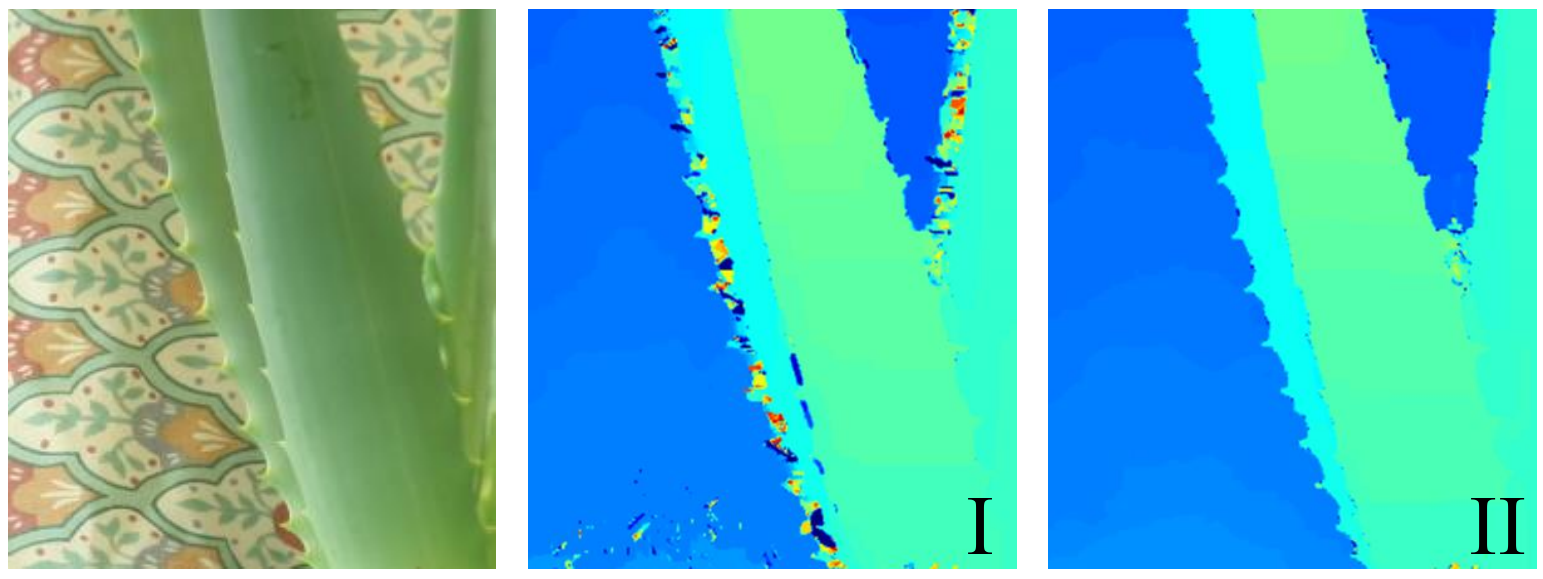}
  \vspace{-0.5cm}
  \caption{Occlusion}
\end{subfigure}
\begin{subfigure}{0.8\columnwidth}
  \centering
  \includegraphics[width=1\columnwidth, trim={0cm 0cm 0cm 0cm}, clip]{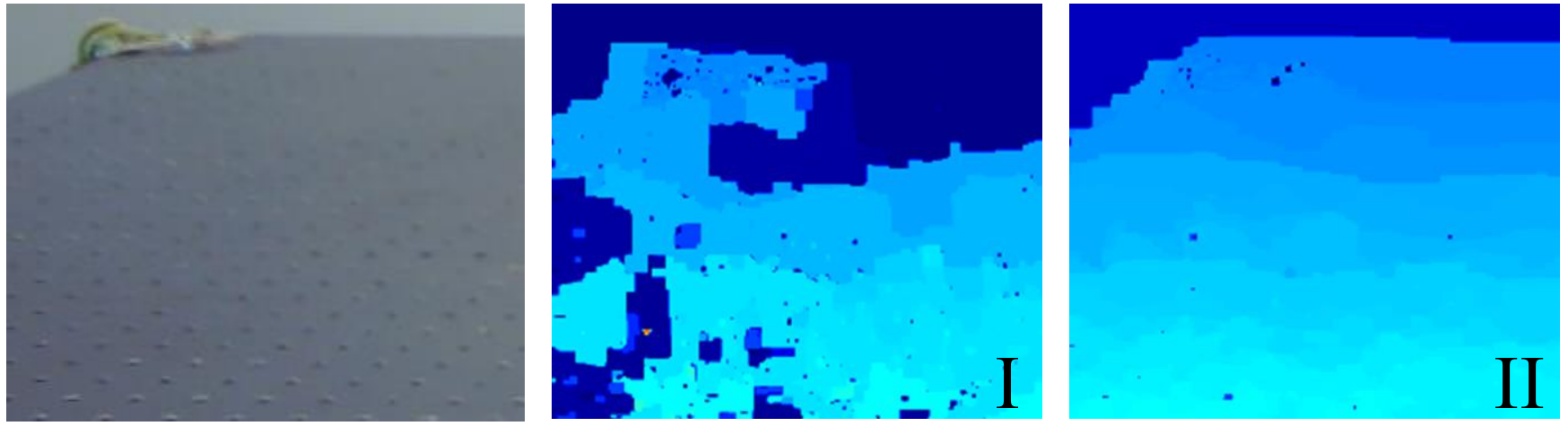}
  \vspace{-0.5cm}
  \caption{Reflection}
\end{subfigure}
\caption{The visual comparison between stereo matching graph cuts (I) and multiscopic graph cuts (II) in noisy, occluded and reflective areas.}
\label{fig:zoom}
\end{figure}

\begin{table*}[]
\centering
\caption{Matching Results of Middlebury Datasets}
\begin{tabular}{c r c c c c c}
\toprule
Data & Methods & RMS & AvgErr & Bad$0.5$ & Bad$1$ & Bad$2$  \\
\midrule
\multirow{6}{*}{Aloe} &
Stereo BM & $4.174$ & $1.021$ & $8.14$ & $5.65$ & $4.32$\\
& Multiscopic BM & $3.176\ (\downarrow 23.9\%)$ & $0.670\ (\downarrow 34.4\%)$ & $5.54$ & $3.62$ & $2.60$\\
& Stereo GC & $3.099$ & $0.731$ & $7.23$ & $4.02$ & $3.27$\\
& Multiscopic GC & $1.416\ (\downarrow 54.3\%)$ & $0.351\ (\downarrow 52.0\%)$ & $4.40$ & $2.15$ & $1.43$\\
& Stereo MC-CNN & $2.527$ & $0.679$ & $5.47$ & $2.63$ & $2.34$\\
& Multiscopic MC-CNN & $1.780\ (\downarrow 29.6\%)$ & $0.469\ (\downarrow 30.9\%)$ & $3.46$ & $1.86$ & $1.72$\\
\midrule
\multirow{6}{*}{Lampshade} &
Stereo BM & $9.534$ & $4.143$ & $24.68$ & $19.87$ & $16.33$\\
& Multiscopic BM & $8.210\ (\downarrow 13.9\%)$ & $3.504\ (\downarrow 15.4\%)$ & $16.50$ & $12.84$ & $10.36$\\
& Stereo GC & $3.088$ & $0.727$ & $6.49$ & $3.90$ & $3.19$\\
& Multiscopic GC & $1.831\ (\downarrow 40.7\%)$ & $0.496\ (\downarrow 31.8\%)$ & $6.27$ & $3.88$ & $2.63$\\
& Stereo MC-CNN & $3.720$ & $0.901$ & $6.61$ & $2.30$ & $1.79$\\
& Multiscopic MC-CNN & $1.796\ (\downarrow 51.7\%)$ & $0.479\ (\downarrow 46.8\%)$ & $4.98$ & $2.05$ & $1.77$\\
\midrule
Average decrease
& Multiscopic GC & $\downarrow 30.5\%$ & $\downarrow 36.2\%$ \\
on all 21 scenes & Multiscopic MC-CNN & $\downarrow \mathbf{58.2\%}$ & $\downarrow \mathbf{50.2\%}$ \\
\bottomrule
\end{tabular}
\label{tab:results}
\end{table*}

\subsection{Evaluation on Middleburry}


The images in the Middlebury Stereo Dataset are well calibrated and rectified, so it can quantitatively show the improvement of multiscopic matching without the influence of image calibration error. Since there are only images captured in the horizontal direction in this dataset, we choose only three images, the view 0, view 1, view 2 as the left image, center image, and right image for the multiscopic vision system. The baseline between view 1 and view 0 or view 2 is 40 mm. Thus with the center image as reference, there are two cost volumes to be combined. One is between the left image and the center image, and the other one is between the right image and the center image. Because these images are rectified, the fusion of these two costs can be directly using the smaller one according to Equ.~\ref{equ:min}.

We randomly choose two sets of images from the Middlebury Stereo Dataset, Aloe and Lampshade, and present their reconstruction results using stereo block matching, stereo graph cuts, multiscopic block matching and multiscopic graph cuts in Fig.~\ref{fig:exp_3frame} without any post-processing. The maximum searching disparity for Aloe and Lampshade is set to $60$ and the minimum is set to $1$. The resolution of these two sets of images are around $1300\times1110$ and the block size for block matching is set to $11$. The occlusion penalty $K$ is set to $10$ and the smoothness parameters $\lambda_1, \lambda_2, \theta, d_{\text{CUTOFF}}$ are set to $9, 3, 8, 5$ respectively. We can visually see from these two sets of results that using three images to do the matching can reduce the noise and reconstruct the occlusion parts better.

For the Middlebury Dataset, we also include a baseline called MC-CNN \cite{zbontar2016stereo} that measures the similarity of image patches with convolutional neural networks, and applies cross-based cost aggregation and semi-global matching. 
The pre-trained accurate Middlebury network model of MS-CNN is used in our experiments. For multiscopic matching, we fuse the cost volume between view 1 and view 0, and the cost volume between view 1 and view 2 according to the minimum rule.

Then we use five metrics to evaluate the matching results, summarized in TABLE \ref{tab:results}. The RMS is the root-mean-square error, AvgErr is the average absolute error, Bad0.5 is the percentage of "bad" pixels whose error is greater than 0.5 and Bad1 and Bad2 denote greater than 1 and 2 respectively. It can be seen from these five metrics that the multiscopic framework can improve the correspondence matching a lot even with only three images. The average decrease on 21 image sequences of the average absolute error can reach $58.2\%$ and the one for root-mean-square error is $50.2\%$. There is around $15\%$ improvement even for the worst case.

\subsection{Real Robot Experiments}

The images captured by our system are not perfectly calibrated and rectified, so there is more noise in the correspondence matching. In our experiments on real robots, we first capture one center image and then capture four surrounding images from the left, right, top and bottom views. The first example, a toy, is presented in Fig.~\ref{fig:stereo} and Fig.~\ref{fig:multi} and another example is presented in Fig.~\ref{fig:exp_5frame}. 

The fusion of four costs is according to the heuristic rule in equation (\ref{equ:heuristic}). The maximum and minimum searching disparity for these two image sets are the same and set to $70$ and $1$. Because the alignment of these data is not perfect, there is more mismatching and noise. Thus the block size for block matching is set to $17$ and the occlusion penalty $K$ is set to $25$ to encourage the correspondence matching. Other parameters are set as the same as Middlebury datasets.

The disparity maps in Fig.~\ref{fig:multi} and Fig.~\ref{fig:exp_5frame} clearly show the multiscopic vision system reduces a lot of noise on texture-less areas, the occlusion parts, and reflective regions. The disparity estimation of reflective metal tabletop is noisy in stereo matching but looks accurate in multiscopic matching, as displayed in Fig.~\ref{fig:zoom}(c). Also, the reflective metal workpiece, which is everywhere in industrial environment, can be reconstructed much better.


\section{CONCLUSION}
\label{sec:conclusion}




In this work, we propose an active perception framework for multiscopic vision. A camera mounted on the end of a robot arm is controlled to move in the image plane and take multiple pictures with the same parallax. Both the magnitude and direction of the pixels disparities is under control such that we can search the correspondence easily. Depth reconstruction with five-frame multiscopic vision is presented in real-world robot experiments. We extend stereo matching algorithms to multiscopic algorithms by fusing four cost volumes between the center frame and surrounding frames, so the outliers in the estimated disparity map could be effectively suppressed. The evaluation on the Middlebury Stereo Dataset and real robot experiments show that a more accurate disparity map could be obtained with our multiscopic vision system. The average absolute error is decreased by $50.2\%$ from stereo matching to multiscopic vision. The noise is significantly reduced on occluded areas and reflective surfaces.

We hope our work with multiscopic vision can inspire more subsequent works in depth estimation and robotic applications. In the future, we can explore the fusion of multiple cost volumes with convolutional neural networks. Also, we can study different image layouts in the multiscopic vision system.




{\small
\bibliographystyle{IEEEtranN}
\bibliography{ref}
}
\end{document}